\documentclass[lettersize,journal]{IEEEtran}
\usepackage{amsmath,amsfonts}
\usepackage[ruled,linesnumbered,vlined]{algorithm2e}
\usepackage{multirow}
\usepackage{amssymb}

\usepackage{array}
\usepackage[caption=false,font=normalsize,labelfont=sf,textfont=sf]{subfig}
\usepackage{textcomp}
\usepackage{stfloats}
\usepackage{url}
\usepackage{verbatim}
\usepackage{graphicx}
\usepackage{cite}
\usepackage{float}

\usepackage{array}

\usepackage{colortbl}
\definecolor{lightgray}{gray}{0.9}
\definecolor{medgray}{gray}{0.8}
\definecolor{midnightblue}{rgb}{0.1, 0.1, 0.64}

\DeclareMathOperator*{\argminA}{arg\,min} 

\let\oldnl\nl
\newcommand{\nonl}{\renewcommand{\nl}{\let\nl\oldnl}}

\begin{document}
\title{PriPHiT: Privacy-Preserving Hierarchical Training of Deep Neural Networks}

\author{Yamin~Sepehri, Pedram~Pad, Pascal~Frossard, and~L.~Andrea~Dunbar
\thanks{Yamin Sepehri and Pascal Frossard are with the Signal Processing Laboratory (LTS4), École
Polytechnique Fédérale de Lausanne (EPFL), CH 1015 Lausanne, Switzerland
(e-mail:yamin.sepehri@epfl.ch; pascal.frossard@epfl.ch).}
\thanks{Yamin Sepehri, Pedram Pad, and L. Andrea Dunbar are with Centre Suisse d'Electronique et de Microtechnique (CSEM), , CH 2002 Neuchâtel, Switzerland
(e-mail: yamin.sepehri@csem.ch; pedram.pad@csem.ch; andrea.dunbar@csem.ch).}
}


\maketitle

\begin{abstract}
The training phase of deep neural networks requires substantial resources and as such is often performed on cloud servers. However, this raises privacy concerns when the training dataset contains sensitive content, e.g., facial or medical images. In this work, we propose a method to perform the training phase of a deep learning model on both an edge device and a cloud server that prevents sensitive content being transmitted to the cloud while retaining the desired information. The proposed privacy-preserving method uses adversarial early exits to suppress the sensitive content at the edge and transmits the task-relevant information to the cloud. This approach incorporates noise addition during the training phase to provide a differential privacy guarantee. We extensively test our method on different facial and medical datasets with diverse attributes using various deep learning architectures, showcasing its outstanding performance. We also demonstrate the effectiveness of privacy preservation through successful defenses against different white-box, deep and GAN-based reconstruction attacks. This approach is designed for resource-constrained edge devices, ensuring minimal memory usage and computational overhead.
\end{abstract}

\vspace{-5pt}
{\section{Introduction}\label{section:introduction}}

\IEEEPARstart{D}{eep} learning models provide state-of-the-art accuracy for various computer vision tasks such as classification, object detection and semantic segmentation~\cite{he2016deep,redmon2016you,ronneberger2015u}. The training phase of deep neural networks often requires significant computational and memory resources as it is a complex task made of forward pass, backward pass and parameter updating and is often done in many iterations on a large number of data samples. Consequently, machines with large computational resources are usually used for training. A popular approach is to use cloud servers provided by service providers. However, the data samples used for training might be private or contain sensitive information such as gender, identity, etc. in addition to the information that the deep neural network should learn. Sharing this information with untrusted cloud service providers increases the concern for privacy attacks~\cite{liu2020privacy,tanuwidjaja2020privacy}. 

A well-established idea is to reduce privacy concerns in deep learning models by using edge-cloud systems. In these systems, the neural network models are divided between the user's device, called the edge, that is trusted and can work with the private data samples, and a large server farm, called the cloud, that has significant computational and memory resources~\cite{teerapittayanon2017distributed,wang2019adda}. In these models, the first few layers of deep neural networks are often implemented at the edge, and the rest of the architecture is deployed in the cloud. As a result, instead of sharing the original input data samples with the cloud, only the intermediate feature maps are communicated. Doing this naturally reduces privacy concerns; however, representation inverters may be able to reconstruct the original inputs using these intermediate feature maps~\cite{dosovitskiy2016inverting,he2019model}.

To preserve the privacy of users against the model inversion attacks, several privacy-preserving approaches have been proposed. Huang et al.~\cite{huang2018generative} proposed an adversarial method to remove the sensitive content from the input samples while keeping the relevant information and Osia et al.~\cite{osia2018deep} proposed the idea of inverse Siamese architectures for the same task. These methods learn to censor the sensitive information from the input while keeping the task-relevant content; however, they are just made for the inference phase. In other words, the training process should be done using enough data samples at the cloud to learn how to preserve privacy and then they can preserve users' privacy in the inference phase on an edge-cloud system. Additionally, although the authors have shown the effectiveness of their methods against different attacks, there is no guarantee that their methods handle the novel attacks that may emerge in the future.

In this study, we propose Privacy-Preserving Hierarchical Training (PriPHiT), a method to train deep neural networks on edge-cloud systems. PriPHiT learns to remove the sensitive content at the edge while keeping the task-relevant information, providing a strong differential privacy guarantee whilst being able to be implemented on resource-constrained edge devices. The model, for the first time, performs adversarial private feature extraction during the training phase using the idea of early exiting~\cite{dosovitskiy2016inverting,sepehri2024hierarchical} at the edge to censor the sensitive content from the feature map that is eventually shared with the cloud, while keeping the task-relevant content. It also benefits from adding noise to the feature map at the edge to maintain a differential privacy guarantee. In contrast to previous works, our framework can train a deep model on an edge-cloud system for the specific use case that is required by the users and the specific sensitive attributes that need to be preserved against privacy attacks. We perform extensive experiments to show the effectiveness of our method to train high-accuracy classifiers on edge-cloud systems using the user's desired content while defending against malicious clouds that train models to classify the user's sensitive content. For example, on a dataset of face data, when the sensitive content is selected to be gender and the desired content is smiling, we show that our method can reduce the accuracy of sensitive content classification attacker to just $0.4\%$ above a blind trivial classifier, while only losing $2.14\%$ of the desired content classification accuracy in comparison to a non-privacy-preserving baseline with the same deep neural network architecture. Moreover, we extensively show the effectiveness of the method against different sorts of inversion attacks such as deep reconstructors and white-box attackers. For instance, in the same smile versus gender experiment on face data, we show that the average SSIM similarity metric between the results of a standard deep reconstruction attacker and the input images is reduced from $0.89$ on a non-privacy-preserving baseline to $0.29$ in PriPHiT model, which indicates an unsuccessful attack according to~\cite{he2019model}. This method can be helpful in the continual training of models on robots that interact with people. For instance, a low-resource camera-holding robot can learn the satisfaction of its users from their facial expressions, without sharing the sensitive content of the users' visual data with the cloud during the learning procedure. In summary, we provide the following contributions:
\begin{itemize}
    \item We propose a new method to hierarchically train deep neural network architectures on edge-cloud systems that preserves the private information of the users in the training dataset. The method provides a differential privacy guarantee for the users, while intelligently keeping the desired information to provide high-accuracy classification in the cloud. 
    \item We show the effectiveness of our privacy-preserving method during the training phase in a scenario when a malicious cloud simultaneously attacks to obtain the sensitive contents and to reconstruct the original inputs using the shared feature maps intended for training our desired content classifier.
    \item We perform extensive experiments on different classification tasks using various neural network architectures to show the performance of our privacy-preserving method to provide high accuracy desired content classification while removing the sensitive content in an unseen test set. More specifically, we show that our privacy-preserving method can reduce the accuracy of sensitive content classification attackers close to blind classifiers while losing a negligible amount of accuracy in the required task.
    \item We provide quantitative and qualitative results showcasing our method's performance against different deep and white-box reconstruction attacks. We show by using the similarity metrics between the outputs of these attacks and the original inputs that their outputs become unrecognizable when they are used against our method.
\end{itemize}

The following parts of the manuscript are structured as follows: In Section~\ref{Sec:related_works}, we discuss works that are related to the different parts of our proposed method. In Section~\ref{Sec:propsoed_method}, we demonstrate our privacy-preserving training approach through the different phases of execution and discuss the attacks during the training phase.  In Section~\ref{Sec:experiments}, we show the results of extensive experiments on different facial and medical datasets with various attributes and the effectiveness of our method against different classification and reconstruction attacks in the training phase and the inference phase. Finally, we conclude our work in Section~\ref{Sec:conclusion}.
\vspace{-12pt}
{\section{Related Works} \label{Sec:related_works}}
In this section, we discuss the works related to our method in two parts: Private feature extraction and preserving privacy in the training phase.
{\subsection{Private Feature Extraction} 
\label{Subsec:private_feature}}
Private feature extraction is a privacy-preserving method in edge-cloud systems whose goal is to find and censor the sensitive content of the input at the edge while keeping the maximum amount of desired information~\cite{guo2023mistnet}. This method has been implemented through two approaches. The first approach proposed by Huang et al.~\cite{huang2017context,huang2018generative} and Li et al.~\cite{li2019deepobfuscator} is by implementing an adversarial game between an edge feature extractor working together with a cloud desired content classifier and a cloud adversary that tries to classify the sensitive content. After finishing this adversarial training, the edge learns to remove the sensitive content while keeping the desired content for the inference phase. The second approach proposed by Osia et al.~\cite{osia2018deep} uses an inverse Siamese network that receives two batches of the data at the edge. It uses a comparison loss on the feature maps extracted by the edge from two input batches. It trains the edge so that if the batches have similar sensitive labels, it pushes away the corresponding output feature maps while doing the opposite for different sensitive labels.

Although these two approaches are powerful against different privacy attacks in the inference phase, they cannot preserve the privacy of users in the training phase so it is not possible to have a private training dataset. Moreover, although they are effective against current privacy attacks, there is no theoretical privacy preservation guarantee in these methods.
\vspace{3pt}
{\subsection{Preserving Privacy in the Training Phase} \label{Subsec:training_privacy}}
Preserving the privacy of users in the training phase is an important topic and it is counted by~\cite{osia2018deep} as a major privacy concern in machine learning. Although it is a less studied topic than the inference phase, it has recently attracted more attention. Yu et al.~\cite{yu2023privacy} implemented a method based on homomorphic encryption to collaboratively train a model on an edge-cloud system. This cryptography-based method encrypts the private data at the edge and collaboratively trains an edge-cloud model that directly works on this encrypted data. The final output is then decrypted by the edge to show the results privately. Although the method provides a strong privacy guarantee for the training phase in edge-cloud systems using blockchain~\cite{gentry2009fully}, it is however too heavy for edge devices that are resource-constrained. Thus, the authors implemented it only on simple models on three simple tabular datasets~\cite{kim2018logistic}. Guo et al.~\cite{guo2023mistnet} proposed another edge-cloud approach that adds noise to the feature map that is extracted by an edge feature extractor before communicating it to the cloud during the training phase. This method also provides a strong differential privacy guarantee~\cite{he2017differential} for the user. However, it is unable to train the edge part of the neural network architecture and this part must be selected from available pre-trained models from similar tasks. Consequently, the selected feature map is not made for the specific task required at the cloud and is not aware of the sensitive information that should be concealed. Moreover, since the feature extractor must be selected from available pre-trained models, there is a question if such a model will exist for novel tasks.

\begin{table*}[hb]
\centering
\caption{Summary of Key Contributions and Limitations in Related Works}
\label{tab:related_works_summary}
\renewcommand{\arraystretch}{1.5} 
\resizebox{\textwidth}{!}{%
\begin{tabular}{|>{\centering\arraybackslash}m{3cm}|>{\centering\arraybackslash}m{4cm}|p{5cm}|p{5cm}|}
\hline
\textbf{Category} & \textbf{Related Work} & \ \ \ \ \ \ \ \ \ \ \ \ \ \ \ \ \textbf{Key Contributions} & \ \ \ \ \ \ \ \ \ \ \ \ \ \ \ \ \ \ \textbf{Limitations} \\ \hline
\multirow{6}{*}{Private Feature Extraction} & \multirow{3}{*}{ Huang et al.~\cite{huang2017context}, Li et al.~\cite{li2019deepobfuscator}} & 
Adversarial training between edge feature extractor, desired classifier, and adversary to remove sensitive content while retaining desired information for the inference phase. &
No privacy guarantee for training phase; effectiveness against specific attacks only; lacks theoretical privacy guarantees. \\ \cline{2-4}
& \multirow{3}{*}{Osia et al.~\cite{osia2018deep}} & 
Uses an inverse Siamese network to push apart feature maps with similar sensitive labels and pull together those with different labels in the inference phase. &
Fails to address training phase privacy; no theoretical privacy guarantees provided. \\ \hline
\multirow{10}{*}{Privacy in Training Phase} & \multirow{2}{*}{Yu et al.~\cite{yu2023privacy}} & 
Homomorphic encryption for collaborative edge-cloud training; provides strong privacy guarantees. &
Resource-heavy for edge devices; applicable only to simple models and datasets. \\ \cline{2-4}
 &  \multirow{2}{*}{Guo et al.~\cite{guo2023mistnet}} & 
Noise added to feature maps for differential privacy; ensures strong privacy for transmitted data. &
Requires pre-trained feature extractors, limiting task-specific performance; sensitive information not explicitly suppressed. \\ \cline{2-4}
 &  \multirow{3}{*}{Gupta and Raskar~\cite{gupta2018distributed}} & 
Distributed training with split neural networks; encryption ensures parameter privacy. &
No protection for transmitted feature maps; requires sharing sensitive labels or multiple communication passes, increasing computational burden and communication costs. \\ \cline{2-4}
&  \multirow{3}{*}{Thapa et al.~\cite{thapa2022splitfed}} & 
Incorporates noise for differential privacy; collaborative training with two servers, one contains a part of the neural network architecture and one for the parameter averaging. &
Noise addition degrades task performance; requires sharing sensitive labels or multiple communication passes, increasing computational burden and communication costs. \\ \hline
\multirow{4}{*}{\textbf{PriPHiT}} & \multirow{3}{*}{---} & 
\textbf{Adversarial early exiting for selective sensitive content suppression; preserves the accuracy of the desired task; avoids label sharing and ensures robust defense against white-box and black-box attacks.} &
 \multirow{3}{*}{\ \ \ \ \ \ \ \ \ \ \ \ \ \ \ \ \ \ \ \ \ \ \ \ \ \ ---}  \\ \hline
\end{tabular}}
\end{table*}

Unlike these approaches, our study proposes a privacy-preserving training method that is suitable for resource-constrained edge devices which intelligently removes the selected sensitive content from the training set while keeping the task-relevant content. Meanwhile, our method provides a strong privacy guarantee. Additionally, in contrast to the previous privacy-preserving training methods such as \cite{guo2023mistnet}, that focus on model inversion and reconstruction attacks and do not work for removing a given sensitive attribute, our method suppresses the sensitive attribute selected by user (e.g., gender), while also preventing reconstruction attacks. Note that this application imposes stricter privacy requirements, as an attacker might be unable to reconstruct the input images but still manage to extract sensitive attributes (e.g., gender).

Another category of related work focuses on Split Federated Learning (SFL), which is a technique for distributed training of neural networks by splitting models between clients and servers. Gupta and Raskar~\cite{gupta2018distributed} introduced an SFL framework where a neural network is split between a cloud server and clients, enabling classification tasks. Their system encrypts and exchanges client parameters but lacks mechanisms to safeguard transmitted feature maps against reconstruction attacks. Thapa et al.~\cite{thapa2022splitfed} extended this work with Splitfed, involving multiple clients and two servers, and introduced noise-based differential privacy for client communicated parameters and feature maps. However, this approach leads to a substantial reduction in model accuracy, even on relatively simple tasks such as MNIST classification~\cite{deng2012mnist}. Both methods require sharing sensitive labels with the server, which is then mitigated by placing final network layers on the client side (in addition to the early layers), however, this substantially increases communication costs and computational demands.

In contrast, our work introduces a privacy-preserving hierarchical training method between an edge device and a cloud server, addressing limitations in existing approaches. By employing a novel adversarial early exiting technique during training, the method selectively retains desired information in transmitted feature maps to maintain task accuracy while eliminating sensitive content, even for complex tasks like facial attribute recognition where the desired information and the sensitive content are heavily mixed. Unlike prior methods, it robustly defends against white-box and advanced deep reconstruction attacks, without sharing sensitive labels or using workarounds that impose significant computational demands or communication costs on edge devices. Table~\ref{tab:related_works_summary} provides a summary of the related works discussed in this section, highlighting their limitations in comparison to our approach.

\begin{figure*}[t!]
\centering
\includegraphics[scale=0.4]{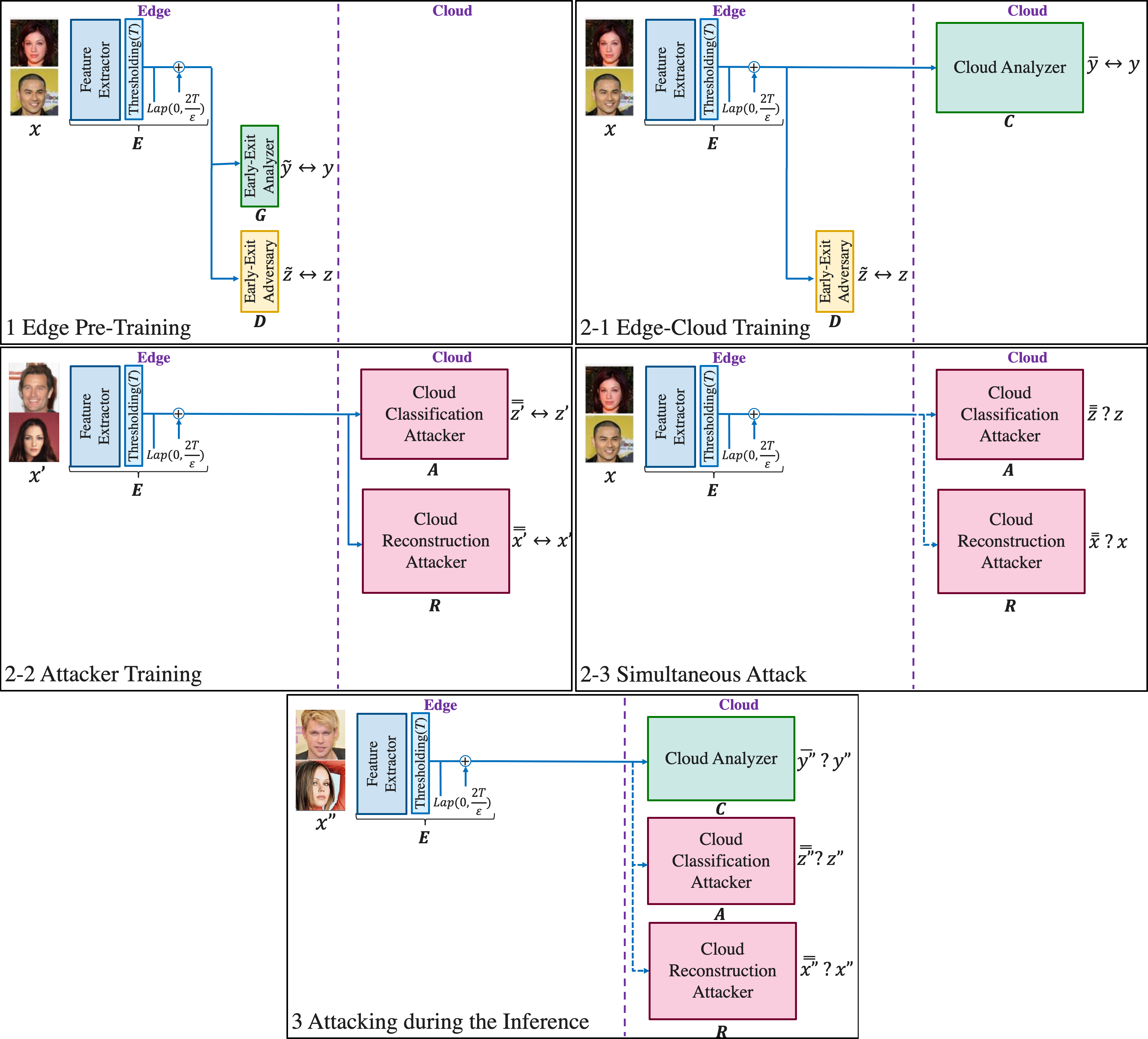}
\caption{The proposed method of privacy-preserving hierarchical training and the steps of edge-cloud execution. Top-left, Step~1: The edge feature extractor is adversarially pre-trained with its two early exits using the user's training set images, the desired content labels and the sensitive content labels. Top-right, Step~2-1: The edge is connected to the cloud and the high-performance cloud analyzer is trained using the sent feature maps extracted from the user's training set images and the desired content labels. The adversarial training of the edge also continues using the early exit. Middle-left, Step~2-2: Since the beginning of the edge-cloud connection, the malicious cloud trains a classification attacker and a reconstruction attacker using its own input images and sensitive content labels at the same time as Step~2-1. Middle-right, Step~2-3: Every time the user sends a feature map from the private training set to train an epoch, the malicious cloud performs a simultaneous attack to classify the sensitive content, or to reconstruct the original input. Bottom, Step~3: When the training is finished, the analyzer and the attackers are tested on a new dataset, unseen by all.}
\label{fig:3-steps}
\end{figure*}

\vspace{2pt}
{\section{Proposed Method} \label{Sec:propsoed_method}}
\vspace{-3pt}

\begin{table}[h]
\centering
\caption{List of Symbols Used in the Proposed Method}
\label{tab:symbols}
\setlength{\tabcolsep}{3pt} 
\renewcommand{\arraystretch}{1.2} 
\footnotesize 
\newcommand{\raiseheight}{\rule[-0.5ex]{0pt}{3.5ex}}
\begin{tabular}{|c|p{6.5cm}|}
\hline
\textbf{Symbol} & \ \ \ \ \ \ \ \ \ \ \ \ \ \  \ \ \ \ \ \ \ \ \textbf{Description} \\ \hline
\( x \) & \raiseheight Input image (e.g., face image). \\ \hline
\( x' \) & \raiseheight Cloud's input dataset for the attackers' training. \\ \hline
\( x_E \) & \raiseheight Feature map extracted by the edge device from \(x\). \\ \hline
\( \overline{\overline{x'}} \) & \raiseheight Reconstruction attacker result in the attackers' training. \\ \hline
\( \overline{\overline{x}} \) & \raiseheight Reconstruction attacker result in the simultaneous attack. \\ \hline
\( \overline{\overline{x''}} \) & \raiseheight Reconstruction attacker result in testing. \\ \hline
\( y \) & \raiseheight Desired content label in the training set. \\ \hline
\( \tilde{y} \) & \raiseheight Predicted desired content by the early-exit analyzer. \\ \hline
\( \bar{y} \) & \raiseheight Predicted desired content by the cloud analyzer in training. \\ \hline
\( y'' \) & \raiseheight Desired content label in the test set. \\ \hline
\( \overline{y''} \) & \raiseheight Predicted desired content by the cloud analyzer in testing. \\ \hline
\( z \) & \raiseheight Sensitive content label in the training set. \\ \hline
\( \tilde{z} \) & \raiseheight Predicted sensitive content by the early-exit adversary. \\ \hline
\( z' \) & \raiseheight Sensitive content label in the cloud's attacker dataset. \\ \hline
\( \overline{\overline{z'}} \) & \raiseheight Predicted sensitive content in the attackers' training. \\ \hline
\( \overline{\overline{z}} \) & \raiseheight Predicted sensitive content in the simultaneous attack. \\ \hline
\( \overline{\overline{z''}} \) & \raiseheight Predicted sensitive content by the classification attacker in testing. \\ \hline
\( \epsilon \) & \raiseheight Privacy budget ensuring differential privacy. \\ \hline
\( T \) & \raiseheight Threshold to bound the feature map. \\ \hline
\( m_a \) & \raiseheight Number of adversarial training steps per epoch. \\ \hline
\( n_p \) & \raiseheight Number of pre-training epochs at the edge. \\ \hline
\( n_t \) & \raiseheight Number of edge-cloud training epochs. \\ \hline
\end{tabular}
\end{table}

In this section, we describe our privacy-preserving hierarchical training approach. Our method is inspired by~\cite{huang2018generative,li2019deepobfuscator} that implemented privacy-preserving methods using an adversarial game between a classifier for the user's desired task and an adversary that classifies the sensitive content. These methods cannot preserve privacy during the training phase. They consider that only the inference phase is done on an edge-cloud system and they need to be trained first to become effective for this privacy-preserving edge-cloud inference. In this study, for the first time, we propose to use the idea of early exiting to keep the adversarial training at the edge. Early exiting is a method in deep architectures that provides a level of decision making in the early layers of models which has been employed in non-privacy-preserving methods to enhance accuracy and reduce runtime~\cite{szegedy2015going,sepehri2024hierarchical}. It is often implemented by additional linear layers that receive the early feature maps. In this work, we propose a step-by-step structure to use early exits to preserve the privacy of users when they want to train a high-accuracy model for a specific task on private datasets and show that the sensitive content is preserved at the users' device (the edge) in all steps.

Moreover, to ensure a theoretical guarantee for privacy protection, we also add noise to the shared information with the cloud during the training. This helps us to ensure differential privacy for the users which is a well-established privacy guarantee. According to the definition of~\cite{Dwork2011}, a random function $f:X\rightarrow Y$ is differentially private with a privacy budget of $\epsilon \geq0$, if for any two inputs $x \in X$ and $\hat{x}\in X$, it satisfies
\begin{equation}
    \label{Eq:diff_privacy}
    \frac{Pr[f(x)=y]}{Pr[f(\hat{x})=y]}\leqslant e^\epsilon
\end{equation}
 for any $y\in Y$. By reducing the privacy budget $\epsilon$, the condition becomes tighter in terms of privacy~\cite{wang2018not}. In this study, we ensure a differential privacy guarantee inspired by a noise-addition mechanism proposed by~\cite{wang2018not} for the inference phase.

Fig.~\ref{fig:3-steps} shows an overview of PriPHiT method and its modules, together with the steps of edge-cloud execution. A complete list of the symbols used in the proposed method can also be found in Table~\ref{tab:symbols}. Our method aims to train a deep architecture within an edge-cloud system to achieve accurate inference of the desired content $y$ from the input image $x$. This must be done in a way that the input image $x$ and its sensitive attribute $z$ are not shared with the cloud. Moreover, the sensitive attribute $z$ must not be extractable from the feature map that is shared with the cloud during the training or inference phases. Additionally, the input image $x$ must not be reconstructable from this shared feature map. This method is made of two steps: edge pre-training and edge-cloud training. In edge pre-training, an edge feature extractor is adversarially pre-trained using two early exits on the user's training set images, the desired content labels and the sensitive content labels. The goal of this step is to make sure that the edge can preserve the user's privacy and remove the sensitive content when it is connected to the cloud for edge-cloud training. In edge-cloud training, the edge is connected to the cloud and a high-performance cloud classifier (the cloud analyzer) is trained for the user's task using the sent feature maps extracted from the training set images and the desired content labels. The adversarial training of the edge also continues using the early exit to make sure that privacy is preserved. Assuming a malicious cloud, in this step, the cloud also begins to attack the shared feature maps using sensitive content classification attackers and input reconstruction attackers. The details of these steps are described in the next parts.

\vspace{2pt}
{\subsection{Edge Pre-Training} \label{subsec:edge-pre-train}}
\vspace{-4pt}
\begin{algorithm}[t!]
    \caption{Edge Pre-Training Procedure}
    \label{alg:edge_pre-training}
    \fontsize{10pt}{11pt}\selectfont

    \KwData{Input training images $x$, desired content labels $y$, sensitive content labels $z$, threshold $T$, privacy budget $\epsilon$, number of pre-training epochs $n_p$, loss coefficient $\lambda$, adversary steps $m_a$}
    \KwResult{pre-trained feature extractor $E$, pre-trained early exit analyzer $G$, pre-trained early exit adversary $D$}

    \BlankLine

    \For{$i \gets 1$ \KwTo $n_p$}{

        $x_{E_1} \gets E_1(x)$ \hfill $\triangleright$ 
        {\footnotesize $E_1$ contains feature extraction layers of $E$}

        $x_{E_2} \gets \frac{x_{E_1}}{\max{(1,\frac{1}{T}\|x_{E_1}\|_\infty)}}$ \hfill $\triangleright$ {\footnotesize Thresholding layer of $E$}

        $x_{E} \gets x_{E_2} + \text{Lap}(0,\frac{2T}{\epsilon})$ \hfill $\triangleright$ {\footnotesize Noise addition block of $E$}

        $\Tilde{y} \gets G(x_{E})$ \hfill $\triangleright$ {\footnotesize Analyzer early exiting}
        
        $\Tilde{z} \gets D(x_{E})$ \hfill $\triangleright$ {\footnotesize Adversary early exiting}
        
        \BlankLine
        $L_{G} \gets [ \sum^K_{k=1} -y_k \log(\Tilde{y}_k)$ \hfill $\triangleright$  \text{\footnotesize Edge analyzer loss}  \\ \nonl $- \lambda (\sum^K_{k=1} -z_k \log(\Tilde{z}_k))  ]$ 
        \BlankLine
        Backward pass from $L_{G}$ and update $E$ and $G$

        \BlankLine
        \For{$j \gets 1$ \KwTo $m_a$}{

        Steps 2-4 to generate $x_{E}$

        $\Tilde{z} \gets D(x_{E})$ \hfill $\triangleright$ {\footnotesize Adversary early exiting}

        $L_{D} \gets \sum^K_{k=1} -z_k \log(\Tilde{z}_k)$ \hfill $\triangleright$ {\footnotesize Edge adversary loss}

        Backward pass from $L_{D}$ and update $D$

        }
    }
    \KwOut{$E$,$D$,$G$}

\end{algorithm}

In this step, the goal is to pre-train an edge feature extractor that extracts features for classification of desired attribute $y$ from the input image $x$, while suppressing the information related to the sensitive attribute $z$. When the pre-training is finished, this feature extractor should be able to provide safe and useful feature maps to train a high-accuracy analyzer in the cloud that classifies the desired attribute. As shown in Fig.~\ref{fig:3-steps}-top-left, the edge receives the private training set from the user that contains input images $x$, desired content labels $y$ and sensitive content labels $z$. For instance, the input images can be images of faces, the desired content label tells if the faces smile and the sensitive content label tells the gender of the person. Algorithm~\ref{alg:edge_pre-training} shows the procedure of edge pre-training. The input image $x$ firstly goes through some layers of the feature extractor, $E$. In addition to a few layers, $E$ contains a thresholding layer that ensures the range of the feature map is less than a specific threshold $T$~\cite{wang2018not}. Then we use the Laplace mechanism~\cite{wang2018not} to add noise to the feature map with a diversity level of $\frac{2T}{\epsilon}$. In this step, i.i.d. Laplace noise is added to all entries of the input. This noise is added to improve the privacy of users by hiding the sensitive content and the diversity level is selected according to the range of the signal to ensure a differential privacy guarantee. Here we prove the differential privacy guarantee of the output feature map of $E$ that is called $x_{E}$ (Line 4 in Algorithm~\ref{alg:edge_pre-training}) with privacy budget $\epsilon$~\cite{wang2018not}:

\begin{equation}
     \frac{Pr[E(x)=x_{E}]}{Pr[E(\hat{x})=x_{E}]} = \frac{Pr[x_{E_2} + \text{Lap}(0,\frac{2T}{\epsilon})=x_{E}]}{Pr[\hat{x}_{E_2} + \text{Lap}(0,\frac{2T}{\epsilon})=x_{E}]}
    \label{eq:edge_feature}
\end{equation}
\begin{equation}
    =\frac{e^{-\frac{|x_{E}-x_{E_2}|\epsilon}{2T}}}{e^{-\frac{|x_{E}-\hat{x}_{{E}_2}|\epsilon}{2T}}}.\nonumber
\end{equation}
We have used a thresholding layer to bound the feature map with a selected threshold $T$~\cite{wang2018not} (Line 3 of Algorithm~\ref{alg:edge_pre-training}), consequently, we have

\begin{equation}
      \frac{e^{-\frac{|x_{E}-x_{E_2}|\epsilon}{2T}}}{e^{-\frac{|x_{E}-\hat{x}_{{E}_2}|\epsilon}{2T}}} =  e^{\frac{|x_{E}-\hat{x}_{{E}_2}|\epsilon}{2T}-\frac{|x_{E}-x_{E_2}|\epsilon}{2T}}  
\end{equation} 

\begin{equation}
 \leqslant e^{\frac{|x_{E_2}-\hat{x}_{{E}_2}|\epsilon}{2T}}\leqslant e^\frac{(2T)\epsilon}{2T}=e^\epsilon. \nonumber
\end{equation} 

As a result, the privacy-preserving feature extractor produces $\epsilon$-differentially private feature maps $x_{E}$. $x_E$ goes to an analyzer early exit that classifies the desired attribute and an adversary early exit that classifies the sensitive content. 

Using the desired labels and sensitive labels, the analyzer loss is produced (Line 6 in Algorithm~\ref{alg:edge_pre-training}). This loss is made of two terms: The first term is a cross-entropy loss that makes sure the feature extractor and the analyzer provide high accuracy for the desired content classification. The second term is a negative cross-entropy loss that makes sure the feature extractor is made in a way that the adversary cannot provide a high-accuracy classification of the sensitive content. There is a $\lambda$ hyperparameter coefficient that regulates the power of these two terms. The next step is the backward pass using the analyzer loss to update the parameters of the feature extractor ($E$) and the analyzer ($G$). In this step, the adversary parameters ($D$) are kept intact. Afterward, for $m_a$ steps, the output of feature map $x_{E}$ is generated and given to the adversary. By comparing the adversary output with sensitive content labels, the adversary loss is generated (Line 11 in Algorithm~\ref{alg:edge_pre-training}) to increase the classification accuracy of the adversary on the sensitive content. Then the backward pass from adversary loss is performed that updates the adversary parameters $D$. The feature extractor $E$ and the analyzer $G$ are kept intact in this step. The adversary update is done for $m_a$ steps to make sure that the adversary is powerful, which results in better performance of the feature extractor to remove the sensitive content. Finally, the whole procedure is done for $n_p$ steps to make sure that the edge is effective enough in removing the sensitive content while keeping the desired content, before connecting it to the cloud for edge-cloud training. $n_p$ is selected according to the speed of the edge device. Higher $n_p$ strengthens the preservation of privacy but has the disadvantage of increasing the computational cost at the edge. 
\begin{algorithm}[t!]
    \caption{Simultaneous Attack Procedure}
    \label{alg:attacker_training}
    \fontsize{10pt}{11pt}\selectfont

    \KwData{Input training images $x'$, sensitive content labels $z'$, number of edge-cloud training epochs $n_t$, feature map extracted from user's private training set $x_{E}$ (different at each epoch)}
    \KwResult{Trained classification attacker $A$, trained reconsturction attacker $R$, classification attacker results from user's communicated feature map $\overline{\overline{z}}$ (different at each epoch), reconstruction attacker results from user's communicated feature map $\overline{\overline{x}}$ (different at each epoch) }

    \BlankLine

    \For{$i \gets 1$ \KwTo $n_t$}{

        $x'_{E} \gets E(x')$ \hfill {\footnotesize $\triangleright$  Edge feature extraction}

        $\overline{\overline{z'}} \gets A(x'_{E})$ \hfill   $\triangleright$ {\footnotesize Classification attack on attacker's data}

        $\overline{\overline{x'}} \gets R(x'_{E})$ \hfill   $\triangleright$ {\footnotesize Reconstruction attack on attacker's data}

        $L_{A} \gets \sum^K_{k=1} -z'_k \log\left(\overline{\overline{z'}}_k\right)$ \hfill $\triangleright$ {\footnotesize Classification attack loss}

        Backward pass from $L_{A}$ and update $A$

        $L_{R} \gets \frac{1}{K}\sum^K_{k=1} 
        (\overline{\overline{x'}}_k-x'_k )^2$ \hfill {\footnotesize$\triangleright$  Reconstruction attack loss}
        
        \BlankLine
        Backward pass from $L_{R}$ and update $R$

        \BlankLine

        $\overline{\overline{z}} \gets A(x_{E})$ \hfill $\triangleright$ {\footnotesize Classification attack on user's feature map}

        $\overline{\overline{x}} \gets R(x_{E})$ \hfill $\triangleright$ {\footnotesize Reconstruction attack on user's feature map}
    }
    \KwOut{$A$,$R$,$\overline{\overline{z}}$,$\overline{\overline{x}}$}

\end{algorithm}
{\subsection{Edge-Cloud Training} \label{subsec:edge-cloud-train}}
When the edge pre-training is finished, the edge device is connected to a cloud server (Fig.~\ref{fig:3-steps}-top-right). The goal of this step is to train a high-accuracy analyzer in the cloud using the extracted features that are sent to the cloud from the edge. We continue using the edge adversary in this step to make sure that the feature extractor continues to suppress the sensitive content effectively. The edge-cloud training procedure is similar to Algorithm~\ref{alg:edge_pre-training} with the only difference of replacing the edge analyzer with the cloud analyzer. This procedure is done for $n_t$ iterations. The value of $n_t$ is selected according to the requirements of the selected deep neural network architecture and it can be similar to the values used by non-privacy-preserving models with a similar architecture.

The cloud system is assumed to be untrusted. As a result, no raw input images $x$ or sensitive content labels $z$ are shared with it. However, we study a scenario that the untrusted cloud uses its own dataset to train attackers to extract the sensitive label or to reconstruct the original input. These attackers use a dataset that contains input images $x'$ and sensitive labels $z'$ to train themselves on the extracted feature maps coming from the edge. Its goal is to learn to classify sensitive content or reconstruct the original inputs from the received feature map. The training procedure is shown in Fig.~\ref{fig:3-steps}-middle-left. In this study, we assume that the cloud classification attacker uses a similar architecture to the cloud analyzer and is trained with the same speed as the analyzer on the cloud. By doing this, after every epoch of classifier attacker training using its own dataset ($x'$ and $z'$), it can perform a simultaneous attack on the feature map that is extracted from the user private input $x$ (that is called $x_E$) and tries to perform a classification on its sensitive content. The simultaneous attack scenario during the edge-cloud training is shown in Fig.~\ref{fig:3-steps}-middle-right.  Algorithm~\ref{alg:attacker_training} shows the training procedure of attackers and their simultaneous attack during the training on the user's feature maps. A complexity analysis of the proposed training method is presented in Supplementary Material~\ref{app:priphit}, which demonstrates that the overall complexity is influenced by factors such as the computational speed of the edge device, the model dimensions, and the number of training samples and steps. Specifically, the edge pre-training phase dominates when the edge device has limited computational capacity or the number of edge pre-training epochs is high. Conversely, the edge-cloud training phase becomes the primary contributor to complexity if the cloud model is significantly large or the number of edge-cloud training epochs is substantial. Detailed runtime analysis and optimization, however, are beyond the scope of this study and are reserved for future work. In the next section, we assess the effectiveness of the proposed hierarchical training method through various experiments.

{\section{Experiments and Results} \label{Sec:experiments}}
In this section, we describe our experiments to evaluate the proposed method. 

\vspace{5pt}
{\subsection{Experimental Settings} \label{Subsec:setting}}
For the experiments, the CelebA dataset~\cite{liu2015faceattributes} is used which is made of facial images of celebrities together with different attributes. We consider two different scenarios where in the first scenario, the gender attribute is considered as the sensitive content, and the smiling attribute is considered as the desired content. In the second scenario, we select wearing makeup as the sensitive content and having a slightly open mouth as the desired content. The dataset is divided into subsets for user training with $90$K samples (which is also used for simultaneous attacks during the training phase),  attacker training with $81$K samples, and testing with $20$K samples. Additionally, the proposed method is also experimented on the FFHQ dataset that is made of face images crawled from Flickr~\cite{karras2019style}, and we used the attributes provided by~\cite{attribute}. As the attribute types are limited for this dataset in comparison to CelebA, this experiment is implemented just for the smiling vs. gender problem. This dataset is also divided into a training subset with $31$K samples (also used for simultaneous attacks), an attacker training subset with $28$K samples, and a testing subset with $10$K samples. 

\begin{table}[t!]
  
  \centering
  \caption{Comparison of the number of parameters as an indicator of memory footprint in the edge and the cloud for PriPHiT in different deep neural network architectures.
  } 
  \resizebox{\columnwidth}{!}{%
  \fontsize{8pt}{11pt}\selectfont
  \begin{tabular}{|c|c|c|c|}
  
    \hline
    
    \multirow{2}{*}{\textbf{Hierarchical Model}} & \multirow{2}{*}{\textbf{System}} & \textbf{No. of parameters}  & \textbf{Maximum PriPHiT Overhead}  \\

     &  & $\mathbf{(\times 10^6)}$ &  \textbf{in Parameters} $\mathbf{(\%)}$ \\

    \hline\hline
    
    \multirow{2}{*}{VGG-11}   & Edge & $0.46$ & $19.53 \%$ \\
    \cline{2-4}
      & Cloud  & $10.17$ & $-$  \\
    \hline\hline

    \multirow{2}{*}{ResNet-18}  & Edge & $0.43$ & $9.05\%$  \\
    \cline{2-4}
      & Cloud & $10.79$ & $-$ \\
    \hline\hline
    \multirow{2}{*}{MobileViT-xxs}  & Edge & $0.30$ & $56.30\%$ \\
    \cline{2-4}
      & Cloud & 17.49 & $-$ \\
     \hline
     
\end{tabular}}
\label{table:param}
\end{table}

\begin{table}[t!]
  
  \centering
  \caption{Comparison of number of the training MACC operations in the edge and the cloud for PriPHiT in different deep neural network architectures.
  } 
  \resizebox{\columnwidth}{!}{%
  \fontsize{8pt}{11pt}\selectfont
  \begin{tabular}{|c|c|c|c|}
  
    \hline
    
    \multirow{2}{*}{\textbf{Hierarchical Model}} & \multirow{2}{*}{\textbf{System}} & \textbf{No. of MACCs}  & \textbf{Maximum PriPHiT Overhead}  \\

    &  &  $\mathbf{(\times 10^6)}$ &  \textbf{in MACCs} $\mathbf{(\%)}$ \\

    \hline\hline
    
    \multirow{2}{*}{VGG-11}   & Edge & $533.39$ & $10.67\%$ \\
    \cline{2-4}
      & Cloud  & $1224.86$ & $-$  \\
    \hline\hline

    \multirow{2}{*}{ResNet-18}  & Edge & $681.00$ & $2.83\%$ \\
    \cline{2-4}
      & Cloud & 1006.48 & $-$\\
    \hline\hline
    \multirow{2}{*}{MobileViT-xxs}  & Edge & 16.38 & $33.44\%$\\
    \cline{2-4}
      & Cloud & 53.18 & $-$\\
     \hline
     
\end{tabular}}
\label{table:mac}
\end{table}

Three popular architectures, VGG-11~\cite{simonyan2014very}, ResNet-18~\cite{he2016deep}, and MobileViT-xxs~\cite{mehta2021mobilevit} are used to test the proposed framework. VGG-11 is divided between the edge and the cloud after the third convolution layer, ResNet-18 is divided after the third residual block, and MobileViT-xxs is divided after the third MobileNet-V2 Block ( Supplementary Material~\ref{app:architecture} ). The edge analyzer and adversary are chosen to be tiny architectures made of one linear layer and one convolution layer. This convolution layer reduces the number of channels and compresses the signal which results in a reduction in the number of neurons needed in the fully connected layer. It helps to reduce the number of parameters in the fully connected layer and the whole edge side which reduces the memory footprint of the edge, making it deployable on low-memory edge devices. It is worth noticing that even though the edge analyzer and adversary are small, the experiments show their significant effectiveness. The number of parameters at the edge and the cloud for the proposed hierarchical training method is shown in Table~\ref{table:param}. According to the table, the number of parameters at the edge is orders of magnitude lower than the cloud. This makes it convenient to run the method on the available memory-constrained edge devices. The number of multiplication and accumulation operations (MACCs) for the training on the edge and on the cloud are also shown in Table~\ref{table:mac}. Similarly, the number of MACCs in this method is lower in the edge in comparison to the cloud, that makes it convenient for slower edge devices. Additionally, according to Tables~\ref{table:param} and~\ref{table:mac}, for both number of parameters and number of MACCs, the overhead of PriPHiT method (for the added modules such as early exits) is small. The only exception is the number of parameters in MobileViT-xxs architecture where half of the parameters belong to the PriPHiT modules, that is due to the tiny number of parameters that this architecture has in the edge as it uses compressed MobileNet-V2 blocks~\cite{sandler2018mobilenetv2}. It is worth mentioning that the we showed the maximum percentage of PriPHiT overhead in Tables~\ref{table:param} and~\ref{table:mac} that is only relevant in the few edge pre-training epochs. These percentages reduce significantly in the edge-cloud training as the analyzer early exit in the edge is not needed anymore. We implemented our model using a powerful GeForce RTX~2080~Ti~\cite{N2080} with $11$ GB of memory (which represents the cloud) and a low-resource Quadro~K620~\cite{k620} with $2$ GB of memory (which represents the edge). In the experiments, we consider that the untrusted cloud designs the reconstruction attacker based on the structure of the edge feature extractor layers according to the suggestions of~\cite{dosovitskiy2016inverting} for inverted model designing. This is a realistic scenario since people often use well-known architectures and the untrusted cloud can guess the structure of the first layers at the edge, based on the last layers implemented in the cloud.

The edge pre-training is done for $n_p=5$ epochs. The number of pre-training epochs is selected to be low to comply with the resource-constrained edge device. Afterward, the edge-cloud training is done for $n_t=50$ epochs. At the same time, the classification and reconstruction attackers are trained for $50$ epochs and perform simultaneous attacks on the feature maps that are shared by the user with the cloud at each epoch. In the next step, the testing phase is done on an unseen subset of the dataset that evaluates the analyzer model, the classification attacker, and the reconstruction attacker. The privacy budget $\epsilon$ is set to $0.5$, $1$, and $2$ for the experiments and the threshold $T$ of the thresholding layer is set to $20$. For the VGG-11 architecture, the loss coefficient is $\lambda = 6$ and the adversary update is done for $m_a=10$ iterations at every epoch. For the ResNet-18 and MobileViT-xxs architectures, we select $\lambda = 9$ and $m_a=15$. To assess the effectiveness of the proposed privacy-preserving hierarchical training method, we propose a baseline that is also an edge-cloud training method aiming to achieve a high-accuracy classification of the desired content at the cloud using similar architectures (VGG-11, Resnet-18, and MobileNet-xxs are split at similar points). Note that the baseline does not use any privacy-preserving methods. Similar to our model, the edge-cloud training of the baseline is done for $50$ epochs and similar classification and reconstruction attackers are trained for $50$ epochs together with this model to execute simultaneous attacks throughout the training process. The baseline analyzer and attackers are also tested on the same unseen dataset at the end, during the inference phase.

{\subsection{Training Phase and Simultaneous Attack} \label{Subsec:training}}
\begin{figure*}[t!]
\centering
\includegraphics[scale=0.5]{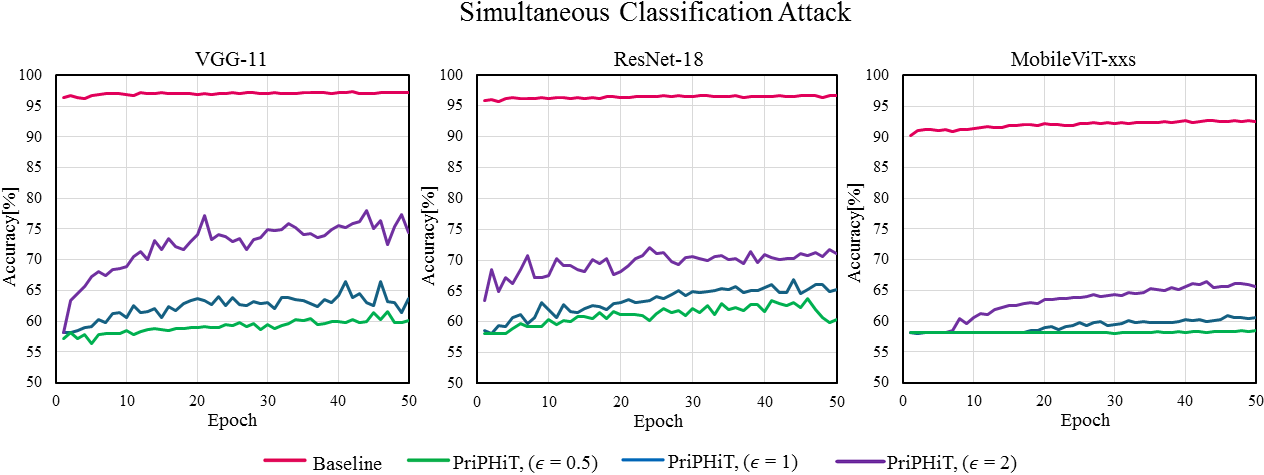}
\caption{The simultaneous classification attack accuracy during the training phase on the feature maps coming from the user's private data at each epoch for the smiling versus gender experiment on the CelebA dataset using different architectures.}
\label{fig:simultaneous_attack_class}
\end{figure*}
\begin{figure*}[t!]
\centering
\includegraphics[scale=0.5]{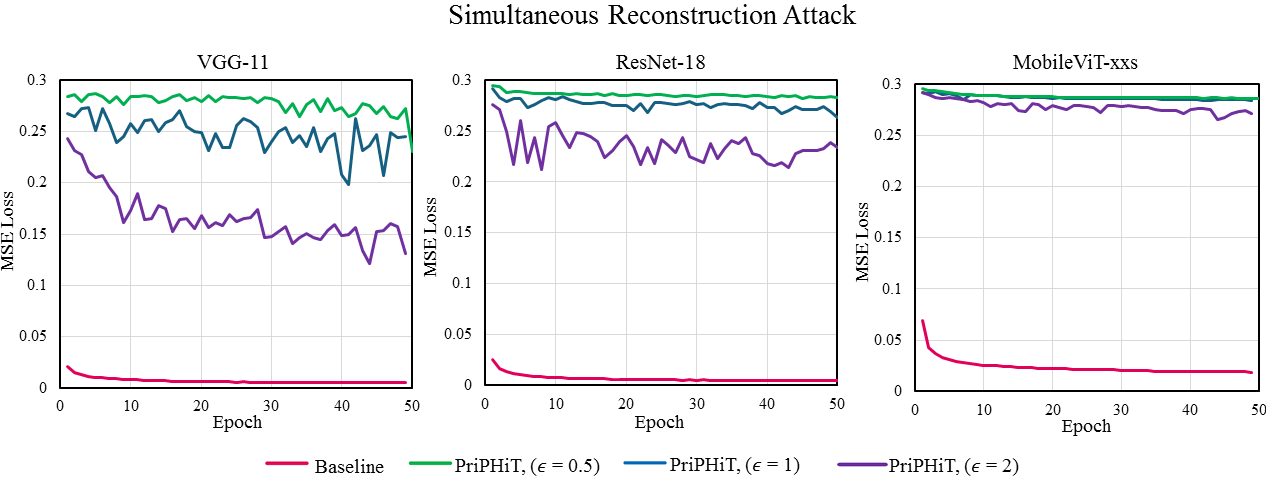} 
\caption{MSE loss of the simultaneous reconstruction attack during the training phase on the feature maps coming from the user's private data compared with the user's original inputs at each epoch for the smiling versus gender experiment on the CelebA dataset using different architectures.}
\label{fig:simultaneous_attack_recon}
\end{figure*}

\begin{figure*}[t!]
\centering
\includegraphics[scale=0.53]{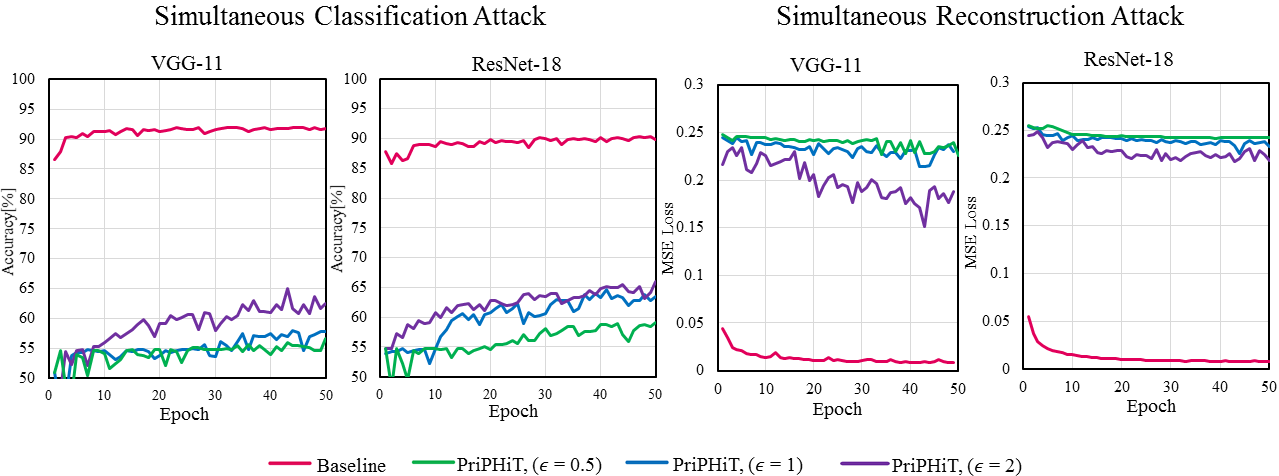}
\caption{The simultaneous attack during the training phase for the smiling versus gender experiment on the FFHQ dataset. Left: The simultaneous classification attack accuracy on the feature maps coming from the user's private data at each epoch. Right: MSE loss of the simultaneous reconstruction attack on the feature maps coming from the user's private data compared with the user's original inputs at each epoch.}
\label{fig:simultaneous_attack_ffhq}
\end{figure*}

Figs.~\ref{fig:simultaneous_attack_class} and~\ref{fig:simultaneous_attack_recon} show the accuracy of the simultaneous classification attack and the mean squared error between the original input samples and the reconstructions performed by the reconstruction attack during the training phase. The results are illustrated for VGG-11, ResNet-18, and MobileViT-xxs architectures with smiling as the desired content and gender as the sensitive content in the experiments on the CelebA dataset. In the baseline model, substantial privacy leakage occurs during the training process, as evidenced by the high-accuracy classification provided by the simultaneously trained classification attacker. However, in the PriPHiT method, the classification attackers fail to provide a high-accuracy classification of the sensitive content from the shared feature maps. Moreover, the trend reaches a plateau, indicating the inability to increase accuracy with more epochs due to the ongoing adversarial training leveraging the edge early exit adversary and the cloud analyzer. Regarding the simultaneous reconstruction attack during the training phase, the baselines exhibit low reconstruction loss with a decreasing trend across epochs, indicating a successful attack. However, in PriPHiT, the attack loss on the user's shared feature map is high and does not change significantly with more epochs, which indicates an unsuccessful simultaneous attack due to the continuous privacy-preserving adversarial training of the feature extractor in the edge.  Supplementary Material~\ref{app:experiment} provides similar plots for experiments involving wearing makeup as the sensitive content and having an open mouth as the desired content.
Additionally, Fig.~\ref{fig:simultaneous_attack_ffhq} shows similar results for smiling versus gender experiments on the FFHQ dataset. Similar to the experiments on the CelebA dataset, we see that our proposed method reduces the accuracy of the simultaneous classification attacker significantly in comparison to the baselines. Additionally, although the average MSE losses of the reconstruction attacks are low for the baselines that indicate a successful attack, the reconstruction losses in the proposed PriPHiT method stay high, indicating unsuccessful attacks. VGG-11 and ResNet-18 architectures are used for the experiments on the FFHQ dataset. Due to the insufficient number of samples in this dataset, implementing and training the MobileViT-xxs architecture is not feasible for this experiment.

\begin{table*}[t!]
  
  \centering
  \caption{Comparison of the test accuracy of PriPHiT with the baselines  for different privacy budgets using various architectures. The results are provided for the two experiments on the CelebA dataset: 1- Desired content: smiling, sensitive content: gender. 2- Desired content: having a slightly open mouth, sensitive content: wearing heavy makeup. For each group of experiments, a trivial classifier shows the lowest accuracy level. It is similar to a blind classifier that just outputs the class with the majority of labels.
  } 
  \scalebox{0.95}{
  \fontsize{8pt}{11pt}\selectfont

  \begin{tabular}{|c|c|c|c|c|}  
    \hline
    
    \multirow{2}{*}{\textbf{Experiment}} & \multirow{2}{*}{\textbf{Architecture}} & \multirow{2}{*}{\textbf{Method}} &\textbf{Analyzer Accuracy} & \textbf{Attacker Accuracy} \\

    &  &  &  \textbf{(desired content)} & \textbf{(sensitive content)} \\

    \hline\hline
       & \multirow{4}{*}{VGG-11} & PriPHiT ($\epsilon=0.5$) & $  88.26\%$ & $ 61.58\%$ \\
       &  & PriPHiT ($\epsilon=1$) & $ 89.80\%$ & $ 64.34\%$ \\
     
     & & PriPHiT ($\epsilon=2$) & $ 90.48\%$ & $ 74.62\%$ \\

      &   & \cellcolor{lightgray} Baseline & \cellcolor{lightgray} $91.16\%$&\cellcolor{lightgray} $97.09\%$ \\
    \cline{2-5}

    \cline{2-5}
    & \multirow{4}{*}{ResNet-18} & 
        PriPHiT ($\epsilon=0.5$) & $88.94\%$ & $61.71\%$ \\
        Smiling&  & PriPHiT ($\epsilon=1$) & $89.03\%$ & $ 65.58\%$ \\
    vs. Gender & & PriPHiT ($\epsilon=2$) & $89.41\%$ & $70.59\%$\\

      &   & \cellcolor{lightgray} Baseline & \cellcolor{lightgray} $91.08\%$ & \cellcolor{lightgray} $96.38\%$ \\
    \cline{2-5}
    & \multirow{4}{*}{MobileViT-xxs} & PriPHiT ($\epsilon=0.5$) & $   86.19\%$ & $  61.52\%$ \\
       &  & PriPHiT ($\epsilon=1$) & $  87.74\%$ & $  62.44\%$ \\
     
     & & PriPHiT ($\epsilon=2$) & $ 88.87\%$ & $  66.37\%$ \\

      &   & \cellcolor{lightgray} Baseline & \cellcolor{lightgray} $ 89.41\%$& \cellcolor{lightgray} $ 92.68\%$ \\
    \cline{2-5}
    \cline{2-5}
     & \multicolumn{2}{c|}{\cellcolor{medgray} Trivial Classifier} & \cellcolor{medgray} $50.01\%$ & \cellcolor{medgray} $61.31\%$ \\
    \hline\hline
        & \multirow{4}{*}{VGG-11} & PriPHiT ($\epsilon=0.5$) & $ 89.76\%$  & $ 59.49\%$ \\
       &  & PriPHiT ($\epsilon=1$) & $ 91.50\%$  & $ 63.31\%$ \\
     & & PriPHiT ($\epsilon=2$) & $ 91.67\%$ & $ 71.76\%$\\

      &   & \cellcolor{lightgray} Baseline & \cellcolor{lightgray} $92.25\%$ & \cellcolor{lightgray} $89.07\%$ \\
    \cline{2-5}
    \cline{2-5}
        & \multirow{4}{*}{ResNet-18} & PriPHiT ($\epsilon=0.5$) & $90.97\%$ & $62.47\%$\\
        Mouth Open&  & PriPHiT ($\epsilon=1$) & $91.48\%$ & $66.32\%$\\
        vs. Makeup& & PriPHiT ($\epsilon=2$) & $ 90.98\%$ & $ 68.52\%$\\

      &   & \cellcolor{lightgray} Baseline & \cellcolor{lightgray} $ 91.83\%$ & \cellcolor{lightgray} $ 87.46\%$ \\
      \cline{2-5}
      \cline{2-5}
    & \multirow{4}{*}{MobileViT-xxs} & 
        PriPHiT ($\epsilon=0.5$) & $ 87.79\%$ & $ 59.55\%$ \\
        &  & PriPHiT ($\epsilon=1$) & $ 89.78\%$ & $  61.24\%$ \\
    & & PriPHiT ($\epsilon=2$) & $ 90.21\%$ & $ 64.41\%$\\

      &   & \cellcolor{lightgray} Baseline & \cellcolor{lightgray} $ 89.48\%$& \cellcolor{lightgray} $ 84.75\%$ \\
    \cline{2-5}
    
    \cline{2-5}
     &  \multicolumn{2}{c|}{\cellcolor{medgray} Trivial Classifier} &\cellcolor{medgray} $ 50.53\%$ & \cellcolor{medgray} $ 59.49\%$\\
    \hline
\end{tabular}
}
\label{table:accuracy}
\end{table*}

\begin{table*}[t!]
  
  \centering
  
  \caption{Comparison of the test accuracy of PriPHiT with the baselines for different privacy budgets using different architectures for the experiments on the FFHQ dataset. The results are provided when the task-relevant content is smiling and the sensitive content is selected to be gender. For the experiments, a trivial classifier shows the lowest accuracy level. It is similar to a blind classifier that just outputs the class with the majority of labels.
  } 
  
  \scalebox{0.95}{
  \fontsize{8pt}{11pt}\selectfont
  \begin{tabular}{|c|c|c|c|c|}  
    \hline
    
    \multirow{2}{*}{\textbf{Experiment}} & \multirow{2}{*}{\textbf{Architecture}} & \multirow{2}{*}{\textbf{Method}} &\textbf{Analyzer Accuracy} & \textbf{Attacker Accuracy} \\

     &  &  &  \textbf{(desired content)} & \textbf{(sensitive content)} \\

    \hline\hline
     \multirow{9}{*}{}  & \multirow{4}{*}{VGG-11} & PriPHiT ($\epsilon=0.5$) & $   82.14\%$ & $  57.82\%$ \\
       &  & PriPHiT ($\epsilon=1$) & $  88.20\%$ & $  58.15\%$ \\
     
     & & PriPHiT ($\epsilon=2$) & $ 87.77\%$ & $  61.30\%$ \\
      Smiling&   & \cellcolor{lightgray} Baseline & \cellcolor{lightgray} $ 92.41\%$& \cellcolor{lightgray} $ 91.54\%$ \\
    \cline{2-5}
    \cline{2-5}
    vs. Gender& \multirow{4}{*}{ResNet-18} & 
        PriPHiT ($\epsilon=0.5$) & $ 88.99\%$ & $ 58.18\%$ \\
        &  & PriPHiT ($\epsilon=1$) & $ 90.21\%$ & $  62.09\%$ \\
    & & PriPHiT ($\epsilon=2$) & $ 90.06\%$ & $ 63.79\%$\\

      &   & \cellcolor{lightgray} Baseline & \cellcolor{lightgray} $ 91.83\%$& \cellcolor{lightgray} $ 89.90\%$ \\
    \cline{2-5}
    \cline{2-5}
     & \multicolumn{2}{c|}{\cellcolor{medgray} Trivial Classifier} & \cellcolor{medgray} $61.31\%$ & \cellcolor{medgray} $56.18\%$ \\
    \hline

\end{tabular}
}
\label{table:accuracy_ffhq}
\end{table*}

{\subsection{Inference Phase and Attacking during the Inference} \label{Subsec:inference_attack}}
After finishing the training of the user's cloud analyzer which is concurrently done with the untrusted cloud's training of the attackers, the user starts the inference phase. We assume that the inference phase is done on a new dataset that is unseen by all models (Fig.~\ref{fig:3-steps}-bottom illustrates the scenario). We assess the efficacy of the cloud analyzer and the edge feature extractor in keeping the desired content by studying the accuracy of the desired content classification. We also test the effectiveness of the edge feature extractor to conceal the sensitive content by studying the accuracy of the trained classification attacker on sensitive content and the similarity metrics between the reconstructed images made by the trained reconstruction attacker and the original inputs. Additionally, we consider another scenario in this phase where the parameters of the trained edge feature extractor are leaked. We assume that the untrusted cloud performs a white-box reconstruction attack using these parameters. For this attack, we use a similar method to the one used in the recent work~\cite{guo2023mistnet}. The details of the white-box reconstruction attack are described in  Supplementary Material~\ref{app:white-box}. We conduct these attacks on all feature maps extracted from the test set and compare them with the original unseen input images quantitatively and qualitatively.

\begin{table*}[!t] 
  
  \centering
  \caption{PSNR and SSIM metrics averaged on the output of the deep reconstruction and white-box reconstruction attacks in comparison to the unseen inputs from the test set for different architectures and pairs of attributes in the experiments on the CelebA dataset.} 
  \fontsize{8pt}{11pt}\selectfont
  \begin{tabular}{|c|c|c|cc|cc|}
  
    \hline
        \multirow{2}{*}{\textbf{Experiment}} & \multirow{2}{*}{\textbf{Architecture}} & \multirow{2}{*}{\textbf{Method}} & \multicolumn{2}{c|}{\textbf{Deep Reconstructor}} & \multicolumn{2}{c|}{\textbf{White-Box Reconstructor}}\\

    & & & \textbf{PSNR} & \textbf{SSIM} & \textbf{PSNR} & \textbf{SSIM} \\
    \hline\hline
      & \multirow{4}{*}{VGG-11} & PriPHiT ($\epsilon=0.5$)& $12.64$& $0.31$ & $7.69$& $0.11$\\
      &  & PriPHiT ($\epsilon=1$)& $12.47$& $0.31$ & $7.72$& $0.14$\\
      & & PriPHiT ($\epsilon=2$) & $14.31$ & $0.38$ & $7.96$ & $0.15$\\

      &   & \cellcolor{lightgray} Baseline & \cellcolor{lightgray} $24.74$ & \cellcolor{lightgray} $0.88$ & \cellcolor{lightgray} $12.02$ &  \cellcolor{lightgray} $0.38$ \\
    \cline{2-7}
        & \multirow{4}{*}{ResNet-18} & PriPHiT ($\epsilon=0.5$)& $11.78$ & $0.29$ & $7.56$ & $0.10$\\
       Smiling&  & PriPHiT ($\epsilon=1$)& $12.12$ & $0.30$ & $7.72$ & $0.14$\\
    vs. Gender& & PriPHiT ($\epsilon=2$) & $12.61$ & $0.31$ & $7.72$ & $0.15$\\

      &   & \cellcolor{lightgray} Baseline & \cellcolor{lightgray} $24.89$ & \cellcolor{lightgray} $0.89$ & \cellcolor{lightgray} $12.16$ & \cellcolor{lightgray} $0.46$\\
    \cline{2-7}
    & \multirow{4}{*}{MobileViT-xxs} & PriPHiT ($\epsilon=0.5$)& $11.74$& $0.28$ & $7.34$& $0.08$\\
       &  & PriPHiT ($\epsilon=1$)& $11.77$& $0.28$ & $ 7.62$& $0.12$\\
     & & PriPHiT ($\epsilon=2$) & $11.91$ & $0.28$ & $7.71$ & $0.14$\\

      &   & \cellcolor{lightgray} Baseline & \cellcolor{lightgray} $20.05$ & \cellcolor{lightgray} $0.74$ & \cellcolor{lightgray} $12.04$ &  \cellcolor{lightgray} $0.46$ \\

    \hline\hline
       & \multirow{4}{*}{VGG-11} & PriPHiT ($\epsilon=0.5$)& $11.81$ & $0.29$ & $7.71$ & $0.12$\\
       &  & PriPHiT ($\epsilon=1$)& $12.22$ & $0.30$ & $7.72$ & $0.14$\\
       & & PriPHiT ($\epsilon=2$) & $13.79$ & $0.36$ & $7.82$ & $0.15$\\

      &   & \cellcolor{lightgray} Baseline & \cellcolor{lightgray} $22.85$ & \cellcolor{lightgray} $0.86$ & \cellcolor{lightgray} $12.22$ & \cellcolor{lightgray} $0.42$ \\
    \cline{2-7}
      & \multirow{4}{*}{ResNet-18} & PriPHiT ($\epsilon=0.5$)& $11.75$ & $0.29$ & $7.75$ & $0.11$\\
       Mouth Open&  & PriPHiT ($\epsilon=1$)& $11.96$ & $0.29$ & $7.70$ & $0.14$\\
    vs. Makeup& & PriPHiT ($\epsilon=2$) & $12.38$ & $0.30$ & $7.72$ & $0.15$\\

      &   & \cellcolor{lightgray} Baseline & \cellcolor{lightgray} $24.62$ & \cellcolor{lightgray} $0.88$ & \cellcolor{lightgray} $11.30$ & \cellcolor{lightgray} $0.40$\\
      \cline{2-7}
      & \multirow{4}{*}{MobileViT-xxs} & PriPHiT ($\epsilon=0.5$)& $11.72$ & $0.28$ & $7.66$ & $0.11$\\
       &  & PriPHiT ($\epsilon=1$)& $11.75$ & $0.28$ & $7.70$ & $0.13$\\
       & & PriPHiT ($\epsilon=2$) & $11.84$ & $0.28$ & $7.69$ & $0.14$\\

      &   & \cellcolor{lightgray} Baseline & \cellcolor{lightgray} $19.91$ & \cellcolor{lightgray} $0.74$ & \cellcolor{lightgray} $10.93$ & \cellcolor{lightgray} $0.43$\\

    \hline
\end{tabular}

 \label{table:metrics}
\end{table*}

\begin{table*}[!t] 
  
  \centering
  
  \caption{PSNR and SSIM metrics averaged on the output of the deep reconstruction and white-box reconstruction attacks in comparison to the unseen inputs from the test subset of the FFHQ dataset for the experiment of smiling versus gender using different architectures.} 
  \fontsize{8pt}{11pt}\selectfont
  \begin{tabular}{|c|c|c|cc|cc|}
  
    \hline
        \multirow{2}{*}{\textbf{Experiment}} & \multirow{2}{*}{\textbf{Architecture}} & \multirow{2}{*}{\textbf{Method}} & \multicolumn{2}{c|}{\textbf{Deep Reconstructor}} & \multicolumn{2}{c|}{\textbf{White-Box Reconstructor}}\\
    & & & \textbf{PSNR} & \textbf{SSIM} & \textbf{PSNR} & \textbf{SSIM} \\
    \hline\hline
      & \multirow{4}{*}{VGG-11} & PriPHiT ($\epsilon=0.5$)& $12.72$& $0.27$ & $8.29$& $0.12$\\
      &  & PriPHiT ($\epsilon=1$)& $13.06$& $0.28$ & $8.27$& $0.14$\\
      & & PriPHiT ($\epsilon=2$) & $13.81$ & $0.31$ & $8.34$ & $0.15$\\
    Smiling&   & \cellcolor{lightgray} Baseline & \cellcolor{lightgray} $21.63$ & \cellcolor{lightgray} $0.84$ & \cellcolor{lightgray} $13.15$ & \cellcolor{lightgray}  $0.44$ \\
    \cline{2-7}
     vs. Gender& \multirow{4}{*}{ResNet-18} & PriPHiT ($\epsilon=0.5$)& $12.43$ & $0.26$ & $8.22$ & $0.11$\\
        &  & PriPHiT ($\epsilon=1$)& $ 12.59$ & $ 0.27$ & $ 8.39$ & $ 0.14$\\
     & & PriPHiT ($\epsilon=2$) & $12.89$ & $0.28$ & $8.30$ & $0.15$\\
      &   & \cellcolor{lightgray} Baseline & \cellcolor{lightgray} $21.98$ & \cellcolor{lightgray} $0.85$ & \cellcolor{lightgray} $11.92$ & \cellcolor{lightgray} $0.36$\\
    \hline

    \hline
\end{tabular}

 \label{table:metrics_ffhq}
\end{table*}

Table~\ref{table:accuracy} shows the accuracy of the VGG-11, ResNet-18, and MobileViT-xxs architectures trained with PriPHiT and the baselines on the unseen test set. The cloud analyzer accuracy in classifying the desired content and the classification attacker accuracy on the sensitive content are shown for two experiments with different pairs of attributes on the CelebA dataset. The analyzer accuracy in the PriPhiT method is close to the baselines, which represents the pinnacle of achievable accuracy. On the other hand, while the baselines exhibit notable privacy leakage, evidenced by the high accuracy of the classification attack on the sensitive content, PriPHiT provides strong preservation of privacy which results in low accuracy in such attacks. When the privacy budget is lower ($\epsilon=0.5$), PriPHiT becomes stronger in removing the sensitive content which costs a slight decrease in terms of the analyzer accuracy in comparison to higher budgets ($\epsilon=2$)\footnote{The effect of $\epsilon$ on the performance of the method is illustrated in Supplementary Material~\ref{app:budget}}. Since the CelebA dataset is not balanced on all attributes, to show the lowest possible accuracy for each experiment, a trivial classifier is also implemented that blindly outputs the majority label. Using the trivial classifier as the minimum and the baseline as the maximum, the effectiveness of the model can be evaluated. For example, in the smiling versus gender experiment using the PriPHiT ($\epsilon=0.5$) method on ResNet-18, PriPHiT can reduce the attacker accuracy down to $61.71\%$ which is just $0.4\%$ higher than the blind trivial classifier. This happens while the analyzer accuracy is $88.94\%$, just $2.14\%$ lower than the maximum accuracy provided by the baseline. Similar accuracy results for the experiment of smiling versus gender on the unseen test subset of the FFHQ dataset are shown in Table~\ref{table:accuracy_ffhq}. Again, it is shown that the PriPHiT method is able to reduce the classification attacker's accuracy on the sensitive content close to the performance level of the blind trivial classifier. Notably, this achievement is attained with only marginal degradation in the analyzer accuracy. Similar to the experiments on the CelebA dataset, when the privacy budget is low (e.g., $\epsilon=0.5$), the attacker's accuracy is reduced more, indicating stronger preservation of privacy, while it costs in lower accuracy of the analyzer.

\begin{figure}[t!]
\centering
\resizebox{\columnwidth}{!}{%
\includegraphics{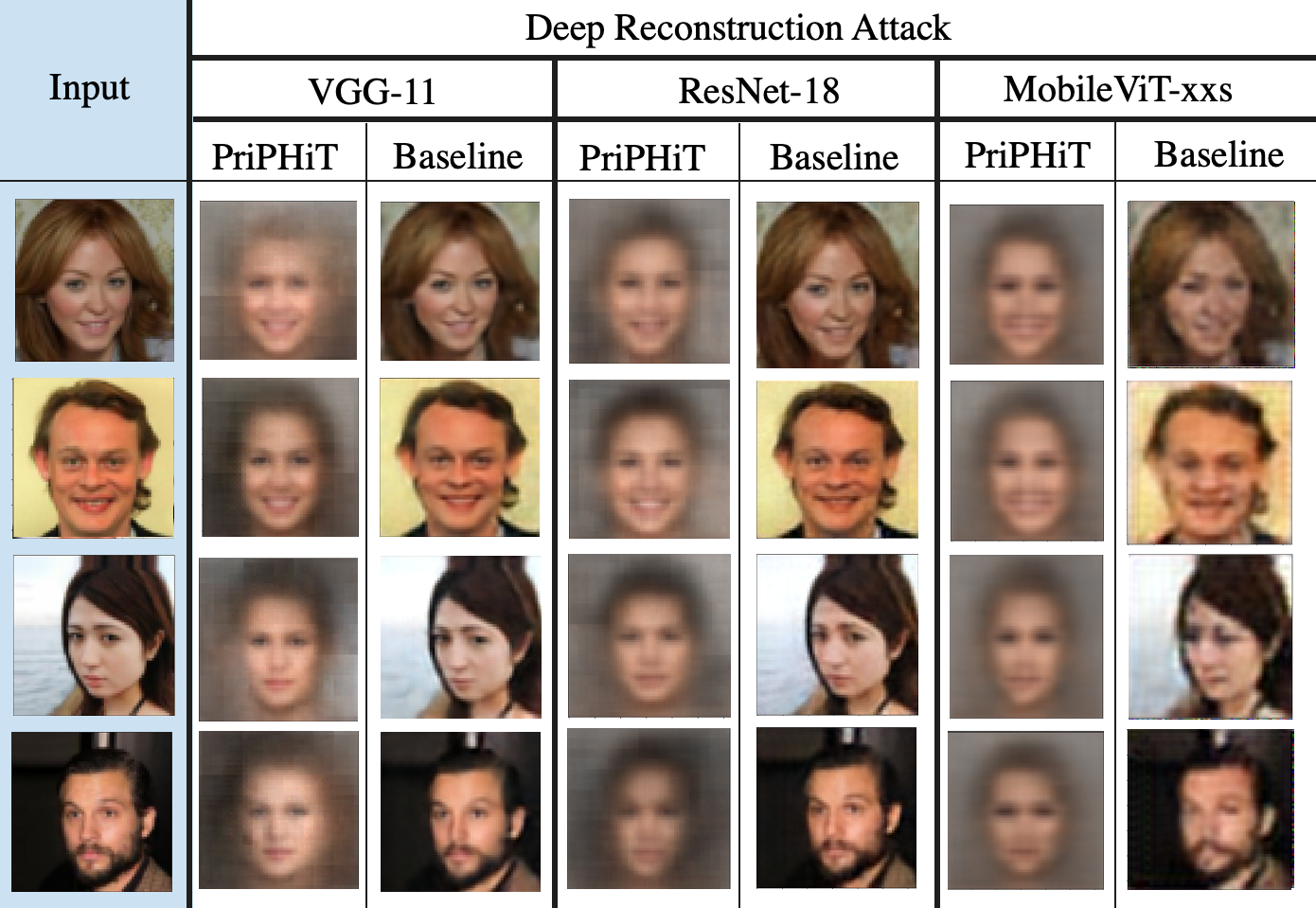}}
\caption{Examples of the results of the deep reconstruction attack on PriPHiT ($\epsilon=1$) in comparison to the similar attacks on the baselines using the unseen inputs from the test subset of the CelebA dataset. \textit{Smiling} and \textit{Gender} attributes are selected as the desired and sensitive contents, respectively. Note the suppression of gender indicators while keeping the smiling attribute in the results of the deep reconstruction attacks when the PriPHiT method is used.}
\label{fig:reconstruction_deep}
\end{figure}

\begin{figure}[t!]
\centering
\resizebox{\columnwidth}{!}{%
\includegraphics{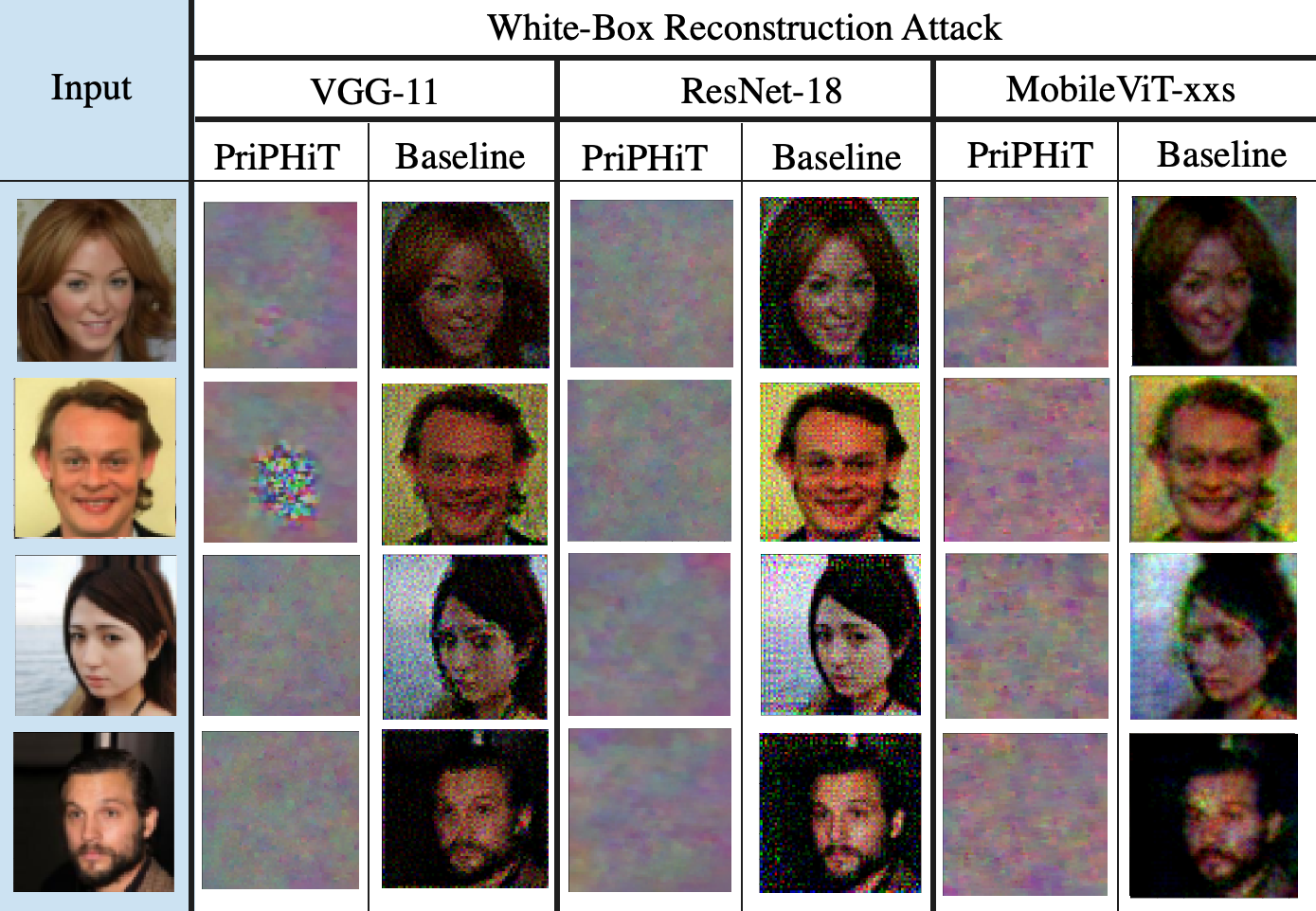}}
\caption{Examples of the results of the white-box reconstruction attack on PriPHiT ($\epsilon=1$) in comparison to the similar attacks on the baselines using the unseen inputs from the test subset of the CelebA dataset. \textit{Smiling} and \textit{Gender} attributes are selected as the desired and sensitive contents, respectively. Note the failure of the attacks when the PriPHiT method is used.}
\label{fig:reconstruction_white}
\end{figure}

\begin{figure}[t!]
\centering
\resizebox{\columnwidth}{!}{%
\includegraphics{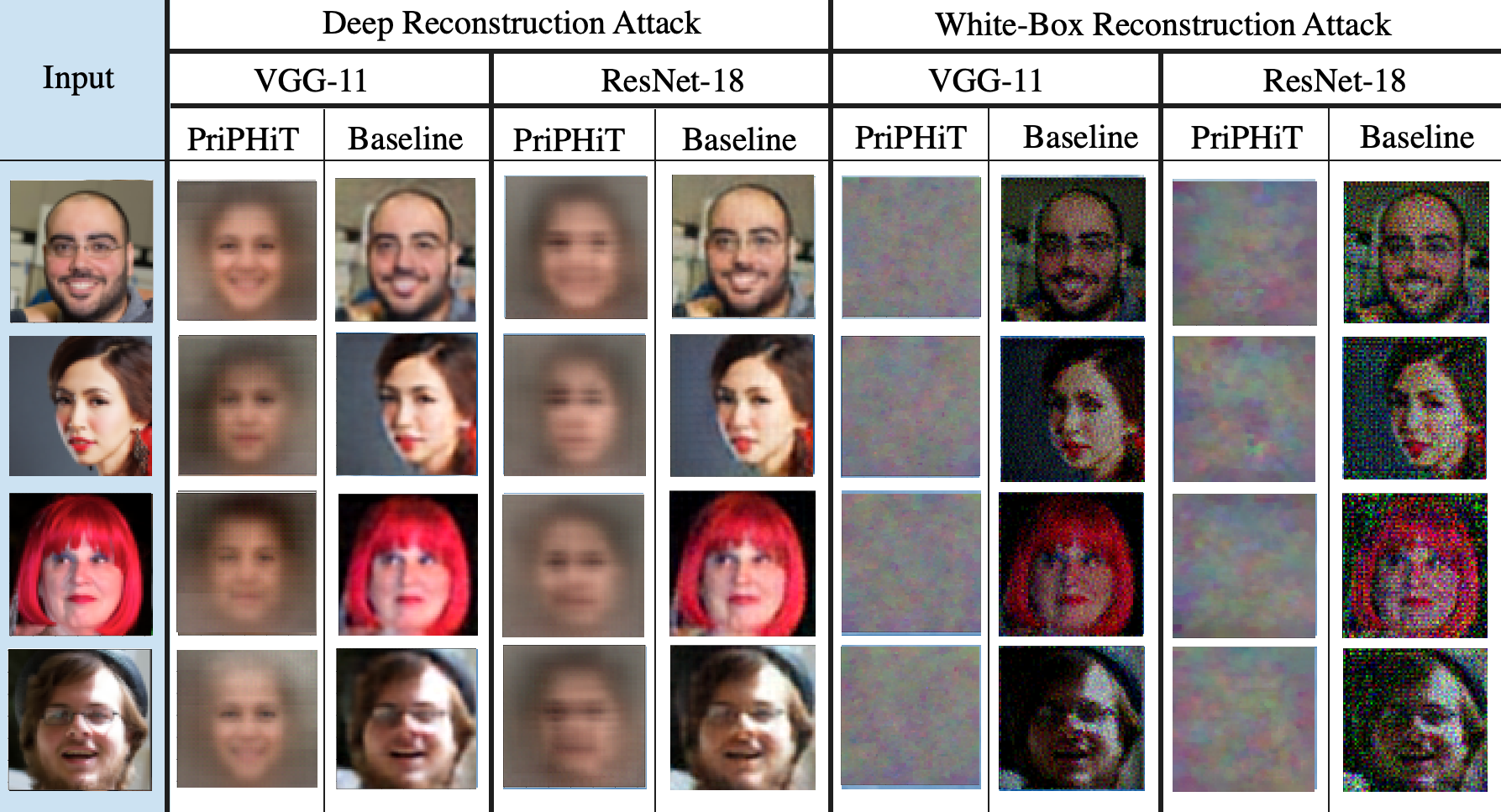}}
\caption{Examples of the results of the deep reconstruction and white-box reconstruction attacks on PriPHiT ($\epsilon=1$) in comparison to the similar attacks on the baselines using the unseen inputs from the test subset of the FFHQ dataset. \textit{Smiling} and \textit{Gender} attributes are selected as the desired and sensitive contents, respectively. Note the suppression of gender indicators while keeping the smiling attribute in the results of deep reconstruction attacks on the PriPHiT method.}
\label{fig:reconstruction_ffhq}
\end{figure}

Table~\ref{table:metrics} shows the PSNR and SSIM metrics on the outputs of the two reconstruction attacks on the PriPHiT method in comparison to the baselines, averaged on the whole test subset for the experiments on the CelebA dataset. Both metrics reduce significantly using the PriPHiT training method while they show high values on the baselines, which indicate successful defense against reconstruction attacks in PriPHiT. Again, lower privacy budgets ($\epsilon=0.5$) provide better defense in comparison to the higher ones ($\epsilon=2$). Examples of the outputs of the reconstruction attacks together with the unseen inputs for the smiling versus gender experiment are shown in Figs.~\ref{fig:reconstruction_deep} and~\ref{fig:reconstruction_white}. We see that the output of both attacks can reconstruct the inputs with gender indicators clearly from the extracted feature maps by the baselines; however, they fail to reconstruct the inputs when the PriPHiT ($\epsilon=1$) method is used. It is interesting to see that in the outputs of the deep reconstruction attacks, the gender indicators (e.g., beard, long hair) are removed while white-box attacks fail completely. Similar figures for the other experiments are shown in  Supplementary Material~\ref{app:qualitative}. Moreover, similar results for the experiments on the FFHQ dataset are observed, as depicted in Table~\ref{table:metrics_ffhq}. Again our results show a notable decrease in the averaged similarity metrics when employing the PriPHiT method, as compared to the baselines. Smaller privacy budgets (e.g., $\epsilon=0.5$) again result in a stronger preservation of privacy. Examples of the outputs of reconstruction attacks for this experiment are shown in Fig.~\ref{fig:reconstruction_ffhq}. It is shown that the deep reconstruction and white-box reconstruction attacks can reconstruct the inputs with gender indicators from the extracted feature maps by the baselines; however, they fail when the PriPHiT ($\epsilon=1$) method is employed.

\begin{figure}[t!]
\centering
\resizebox{\columnwidth}{!}{%
\includegraphics{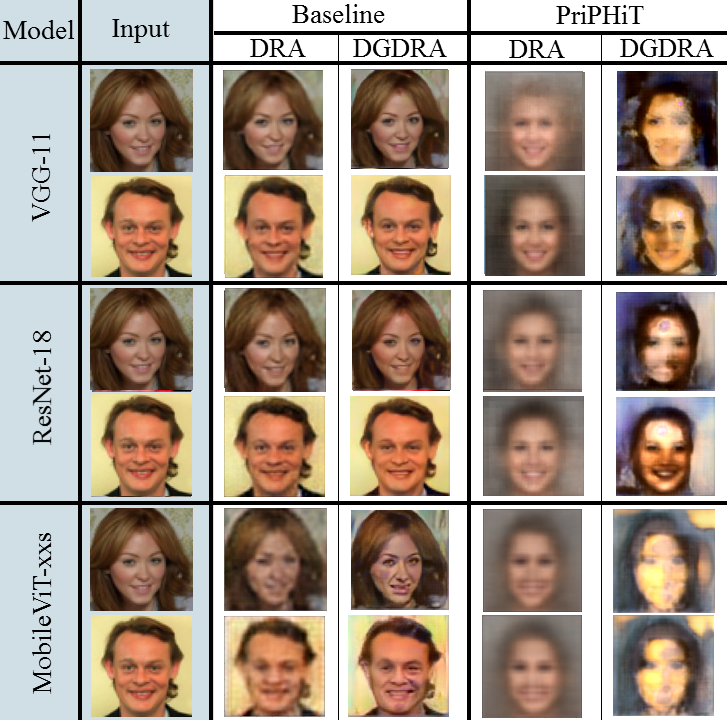}}
\caption{Examples of the results of the original Deep Reconstruction Attack (shown as DRA) and DeblurGAN-v2-boosted Deep Reconstruction Attack (shown as DGDRA) on PriPHiT ($\epsilon=1$) in comparison to the similar attacks on the baselines using the unseen inputs from the test subset of the CelebA dataset. \textit{Smiling} and \textit{Gender} attributes are selected as the desired and sensitive contents, respectively. Note that the DeblurGAN-v2-Boosted attack makes sharper and more clear reconstruction on Baseline model in comparison to the original Deep Reconstruction Attack, while it fails to improve the results when our PriPHiT method is used.}
\label{fig:deblurganv2}
\end{figure}

To demonstrate the effectiveness of our privacy-preserving method against more advanced deep reconstruction attacks, we evaluate a more sophisticated attack by training a DeblurGAN-v2 architecture~\cite{kupyn2019deblurgan} on top of our reconstruction attacker, which was designed based on the suggestions of~\cite{dosovitskiy2016inverting}. The results of this experiment, conducted on the CelebA dataset, are presented in Table \ref{tab:deblur} and Figure \ref{fig:deblurganv2}. As indicated by the results, this reconstruction attack leads to sharper and more precise outputs for the baseline models across different architectures, as evidenced by higher PSNR values in Table \ref{tab:deblur} and clearer visual results in Figure \ref{fig:deblurganv2}. However, when the PriPHiT method is applied, the attack fails to improve the quality of the reconstructed images, highlighting the robustness of our method against GAN-enhanced attacks.

\vspace{5pt}

{\subsection{Experiments on Medical Data} \label{Subsec:medical}}

To further demonstrate the effectiveness of our privacy-preserving hierarchical training method on private data types beyond facial images, we conducted experiments on medical data samples. Specifically, we trained similar deep neural network architectures using PriPHiT on the CheXpert dataset~\cite{irvin2019chexpert}, which consists of chest X-ray images of patients. For these experiments, we used a subset of CheXpert provided by~\cite{chexpertsmall}. We selected having symptoms as the desired attribute and the patient’s gender as the sensitive attribute. Due to a significant class imbalance in the dataset—where symptomatic samples far outnumber benign samples—we balanced the dataset by randomly subsampling symptomatic images to match the number of benign images. The data was then divided into $20$K samples for training (also used for simultaneous attacks during the training phase), $18$K samples for attacker training, and approximately $7$K samples for testing. Similar to the experiment on the FFHQ dataset, VGG-11 and ResNet-18 architectures were used for the experiments on the CheXpert dataset, as the number of samples was not sufficient for training the MobileViT-xxs architecture.

\begin{figure*}[t!]
\centering
\resizebox{\textwidth}{!}{
\includegraphics{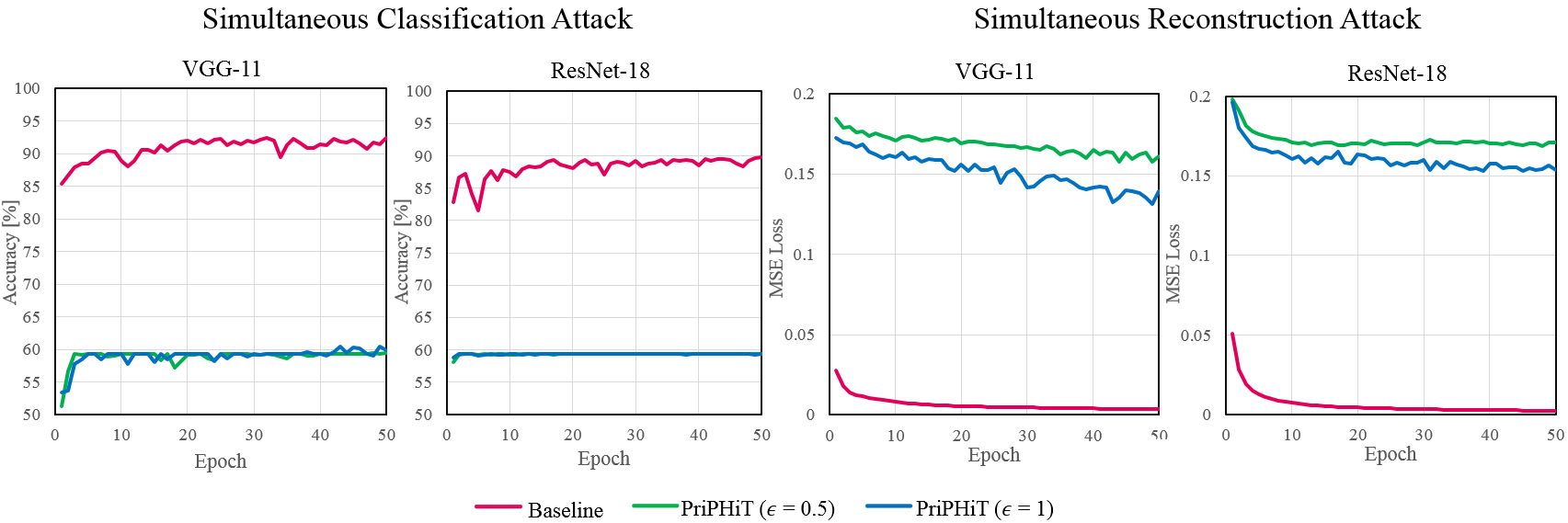}}
\caption{The simultaneous attack during the training phase for the having symptom versus gender experiment on the CheXpert dataset. Left: The simultaneous classification attack accuracy on the feature maps coming from the user's private data at each epoch. Right: MSE loss of the simultaneous reconstruction attack on the feature maps coming from the user's private data compared with the user's original inputs at each epoch.}
\label{fig:chex_simultanous}
\end{figure*}

Figure~\ref{fig:chex_simultanous} illustrates the results of simultaneous classification and reconstruction attacks over 50 training epochs when our PriPHiT method is applied, compared to baselines with similar architectures but without privacy-preserving features. In the baseline models, the accuracy of simultaneous classification attacks is high and continues to increase with more epochs, indicating a successful attack on the users’ sensitive content. In contrast, when PriPHiT is used, the accuracy of classification attacks remains low and close to that of a trivial classifier that blindly predicts the majority label, demonstrating PriPHiT's effectiveness in preserving privacy during the training phase. Furthermore, the MSE Loss between the outputs of simultaneous reconstruction attackers and the inputs in the baseline models approaches zero, indicating successful reconstruction attacks. However, when PriPHiT is employed, the MSE Loss remains high and plateaus, further illustrating PriPHiT's ability to defend against simultaneous reconstruction attacks.

\begin{table*}[t!]
\centering
\caption{Performance of the PriPHIT and baseline methods against the DeblurGAN-V2-boosted deep reconstruction attacker~\cite{kupyn2019deblurgan} in the smiling versus gender experiment on the CelebA dataset.}
\label{tab:deblur}
\begin{tabular}{|c|c|cc|cc|}
\hline
\multirow{3}{*}{\textbf{Architecture}}           & \multirow{3}{*}{\textbf{Model}} & \multicolumn{2}{c|}{\textbf{Deep Reconstruction Attack}} & \multicolumn{2}{c|}{\textbf{DeblurGANv2-Boosted }} \\ 

          & &  & & \multicolumn{2}{c|}{\textbf{Deep Reconstruction Attack}} \\ 

                         &           & \textbf{PSNR}       & \textbf{SSIM}     & \textbf{PSNR}       & \textbf{SSIM}     \\ 
\hline
\hline
\multirow{3}{*}{VGG-11}  & PriPHIT ($\epsilon$=0.5) & 12.64             & 0.31           & 10.17             & 0.38           \\ 
                         & PriPHIT ($\epsilon$=1)   & 12.47             & 0.31           & 10.75             & 0.35           \\ 
                         & \cellcolor{lightgray} Baseline   & \cellcolor{lightgray} 24.74             & \cellcolor{lightgray} 0.88           & \cellcolor{lightgray} 28.29             & \cellcolor{lightgray} 0.88           \\ 
\hline
\multirow{3}{*}{ResNet-18} & PriPHIT ($\epsilon$=0.5) & 11.78             & 0.29           & 9.88              & 0.37           \\ 
                         & PriPHIT ($\epsilon$=1)   & 12.12             & 0.30           & 10.05             & 0.36           \\ 
                         &\cellcolor{lightgray}  Baseline   &\cellcolor{lightgray}  24.89             & \cellcolor{lightgray} 0.89           & \cellcolor{lightgray} 27.91             & \cellcolor{lightgray} 0.90           \\ 
\hline
\multirow{3}{*}{MobileViT-xxs} & PriPHIT ($\epsilon$=0.5) & 11.74             & 0.28           & 10.54             & 0.34           \\ 
                         & PriPHIT ($\epsilon$=1)   & 11.77             & 0.28           & 10.33             & 0.36           \\ 
                         & \cellcolor{lightgray} Baseline   & \cellcolor{lightgray} 20.05             & \cellcolor{lightgray} 0.74           & \cellcolor{lightgray} 22.62             & \cellcolor{lightgray} 0.75           \\ 
\hline
\end{tabular}
\end{table*}

After completing the training phase, the models are evaluated on the unseen test set. The accuracy results of the analyzer and attacker models during the inference phase are presented in Table \ref{table:accuracy_chexpert}. Similar to the previous experiments, the classification attack accuracy on PriPHiT is significantly lower compared to the baseline models, indicating the effectiveness of our method in removing sensitive content from the shared feature map. This accuracy is very close to that of a trivial classifier. Conversely, the accuracy of the analyzer model on the desired content remains high and comparable to the baseline when PriPHiT is applied, demonstrating that PriPHiT preserves task-relevant information. Notably, in some experiments, the analyzer accuracy slightly exceeds that of the baseline, which may be attributed to a regularization effect introduced by the added noise~\cite{nagel2021white}.

\begin{figure*}[t!]
\centering
\includegraphics[scale=0.5]{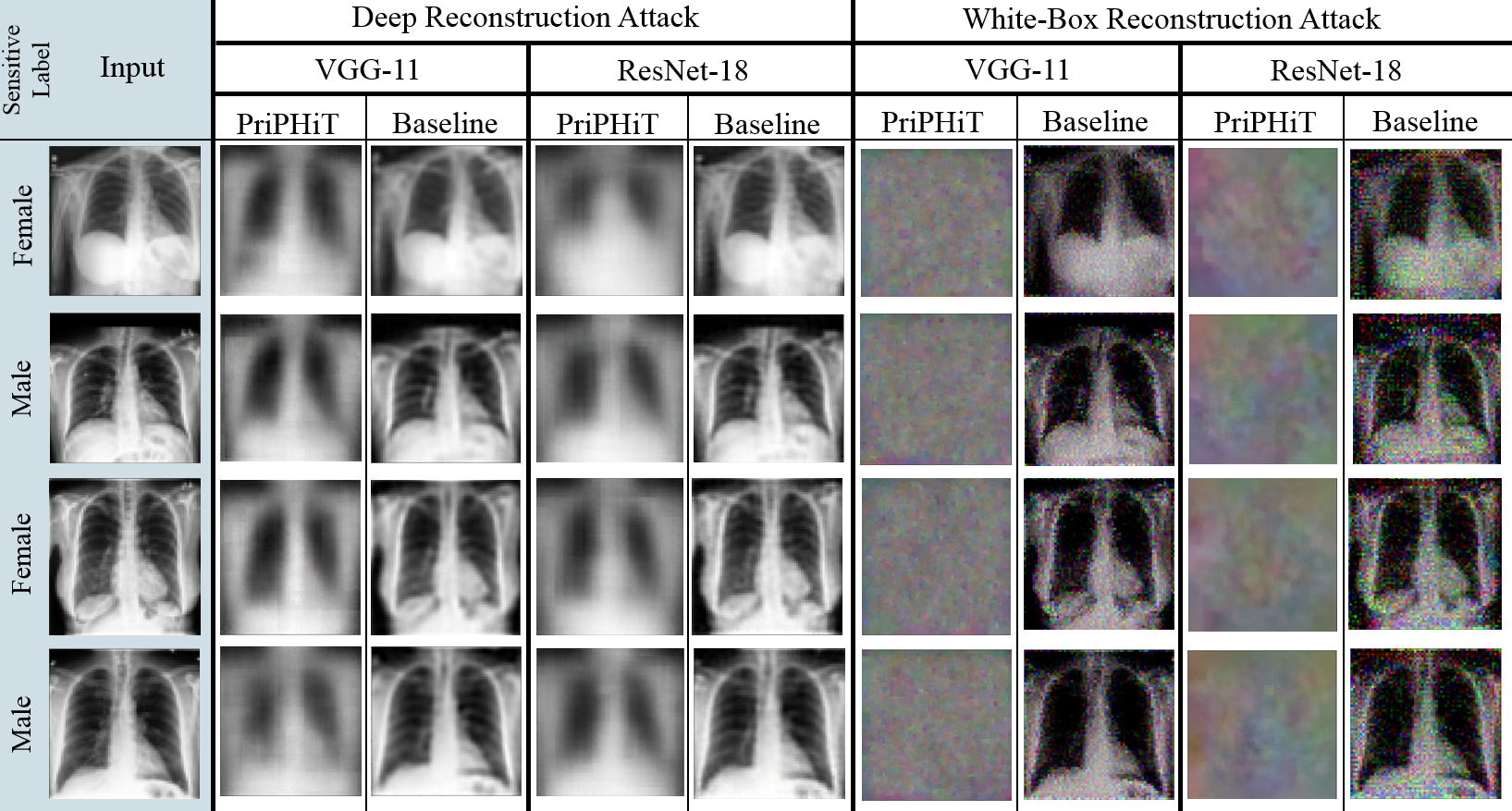}
\caption{Examples of the results of the deep reconstruction and white-box reconstruction attacks on PriPHiT ($\epsilon=1$) in comparison to the similar attacks on the baselines using the unseen inputs from the test subset of the CheXpert dataset. \textit{Having Symptom} and \textit{Gender} attributes are selected as the desired and sensitive contents, respectively. Note the suppression of gender indicators (e.g., breasts) in the results of deep reconstruction attacks and the failure of white-box reconstruction attacks when the PriPHiT method is used.}
\label{fig:reconstruction_chexpert}
\end{figure*}

The similarity metrics for the outputs of reconstruction attacks are shown in Table \ref{tab:recon_chexpert}, with examples of the outputs from deep reconstruction and white-box reconstruction attacks provided in Figure \ref{fig:reconstruction_chexpert}. As shown in the table, the similarity metrics are significantly lower when the PriPHiT method is applied, particularly with tighter privacy budget values ($\epsilon=0.5$), compared to the baselines. This demonstrates the effectiveness of our method in defending against these attacks. In the baseline models, deep reconstruction attackers successfully generate clear reconstructions of the original unseen inputs, including sensitive gender indicators (e.g., breasts). However, when PriPHiT is used, the deep reconstruction attacker fails to reconstruct these sensitive features. Similarly, white-box reconstruction attacks succeed in the baseline models, producing inputs containing sensitive content, but fail entirely when PriPHiT is employed.

\begin{table*}[t!]
  
  \centering
  
  \caption{Comparison of the test accuracy of PriPHiT with the baselines for different privacy budgets using different architectures for the experiments on the CheXpert dataset. The results are provided when the task-relevant content is having symptoms and the sensitive content is selected to be gender. For the experiments, a trivial classifier shows the lowest accuracy level. It is similar to a blind classifier that just outputs the class with the majority of labels.
  } 
  \fontsize{8pt}{11pt}\selectfont
  \begin{tabular}{|c|c|c|c|c|}  
    \hline
    
    \multirow{2}{*}{\textbf{Experiment}} & \multirow{2}{*}{\textbf{Architecture}} & \multirow{2}{*}{\textbf{Method}} &\textbf{Analyzer Accuracy} & \textbf{Attacker Accuracy} \\

     &  &  &  \textbf{(desired content)} & \textbf{(sensitive content)} \\

    \hline\hline
     \multirow{7}{*}{}  & \multirow{3}{*}{VGG-11} & PriPHiT ($\epsilon=0.5$) & $   73.38\%$ & $  59.16\%$ \\
       &  & PriPHiT ($\epsilon=1$) & $  74.58\%$ & $  59.86\%$ \\
     
      Smiling&   & \cellcolor{lightgray} Baseline & \cellcolor{lightgray} $ 72.22\%$& \cellcolor{lightgray} $ 92.05\%$ \\
    \cline{2-5}
    \cline{2-5}
    vs. Gender& \multirow{3}{*}{ResNet-18} & 
        PriPHiT ($\epsilon=0.5$) & $ 71.20\%$ & $ 59.18\%$ \\
        &  & PriPHiT ($\epsilon=1$) & $ 70.33\%$ & $  59.18\%$ \\

      &   & \cellcolor{lightgray} Baseline & \cellcolor{lightgray} $ 70.76\%$& \cellcolor{lightgray} $ 89.89\%$ \\
    \cline{2-5}
    \cline{2-5}
     & \multicolumn{2}{c|}{\cellcolor{medgray} Trivial Classifier} & \cellcolor{medgray} $50.00\%$ & \cellcolor{medgray} $59.14\%$ \\
    \hline

\end{tabular}
\label{table:accuracy_chexpert}
\end{table*}

\begin{table*}[!t] 
  
  \centering
  
  \caption{PSNR and SSIM metrics averaged on the output of the deep reconstruction and white-box reconstruction attacks in comparison to the unseen inputs from the test subset of the CheXpert dataset for the experiment of having symptom versus gender using different architectures.} 
  \fontsize{8pt}{11pt}\selectfont
  \begin{tabular}{|c|c|c|cc|cc|}
  
    \hline
        \multirow{2}{*}{\textbf{Experiment}} & \multirow{2}{*}{\textbf{Architecture}} & \multirow{2}{*}{\textbf{Method}} & \multicolumn{2}{c|}{\textbf{Deep Reconstructor}} & \multicolumn{2}{c|}{\textbf{White-Box Reconstructor}}\\
    & & & \textbf{PSNR} & \textbf{SSIM} & \textbf{PSNR} & \textbf{SSIM} \\
    \hline\hline
      & \multirow{3}{*}{VGG-11} & PriPHiT ($\epsilon=0.5$)& $ 14.21$& $ 0.39$ & $ 8.97$& $ 0.13$\\
      &  & PriPHiT ($\epsilon=1$)& $ 14.81 $& $ 0.43$ & $ 9.16$& $ 0.15$\\
    Smiling&   & \cellcolor{lightgray} Baseline & \cellcolor{lightgray} $ 25.59$ & \cellcolor{lightgray} $ 0.92$ & \cellcolor{lightgray} $ 14.02$ & \cellcolor{lightgray}  $ 0.55$ \\
    \cline{2-7}
     vs. Gender& \multirow{3}{*}{ResNet-18} & PriPHiT ($\epsilon=0.5$)& $ 13.99$ & $ 0.37$ & $ 9.23$ & $ 0.13$\\
        &  & PriPHiT ($\epsilon=1$)& $ 14.45$ & $ 0.40$ & $ 9.17$ & $ 0.16$\\
      &   & \cellcolor{lightgray} Baseline & \cellcolor{lightgray} $ 21.59$ & \cellcolor{lightgray} $ 0.92$ & \cellcolor{lightgray} $ 13.54$ & \cellcolor{lightgray} $ 0.46$\\
    \hline

    \hline
\end{tabular}

 \label{tab:recon_chexpert}
\end{table*}

\vspace{5pt}

{\section{Conclusion and Future Works} 
\label{Sec:conclusion}}
In this study, we proposed PriPHiT, a privacy-preserving method to hierarchically train deep neural networks on edge-cloud systems. The method leverages the idea of early exit to preserve the privacy of users in the training phase for the first time and provides a feature extraction that removes the selected sensitive content from the training dataset while keeping the task-relevant information which is then classified at the cloud. The outstanding performance of the proposed method has been demonstrated through extensive experiments on diverse pairs of sensitive content and desired content from different datasets using various neural network architectures. We showed that our method provides a differential privacy guarantee and effectively prevents deep reconstruction attacks and white-box reconstruction attacks, as evidenced by quantitative and qualitative results. 

A future line of research is to study how the computational resources available at the edge impact both privacy preservation and runtime. Given that our approach involves minimal additions to the edge (such as early exits with limited parameters), exploring its effect on edge power consumption could be valuable, as the edge devices may have confined energy sources. Additionally, as there is no technical barrier to scale the proposed method to be implemented on larger datasets except the total training runtime, a runtime analysis using a set of different available edge devices and datasets is practically valuable.

 

\twocolumn[\begin{@twocolumnfalse}\textit{
\textbf{Supplementary Material for PriPHiT: Privacy-Preserving Hierarchical Training of Deep Neural Networks}}

\textit{\textbf{By Yamin Sepehri, Pedram Pad, Pascal Frossard, and L. Andrea Dunbar.}}
\end{@twocolumnfalse}]

\ 
\setcounter{section}{0}
\renewcommand{\thesection}{\Alph{section}}  

{\section{Complexity Analysis of PriPHiT} \label{app:priphit}}

In this section, we perform a simplified complexity analysis of the PriPHiT method. For simplicity, we assume that the communication between edge and cloud takes a negligible time. We divide this analysis into the two phases of edge pre-training and edge-cloud training (see Sections \ref{subsec:edge-pre-train} and \ref{subsec:edge-cloud-train}).

\subsection*{1. Edge Pre-Training (Section~\ref{subsec:edge-pre-train})}
\begin{itemize}
    \item \textbf{Feature Extractor:} Feature extraction involves forward passes and backward passes through the edge feature extractor, which typically has complexity:
    \begin{equation}
    O(m \cdot n^2 \cdot k^2),
    \end{equation}
    where $m$ is the number of filters, $n$ is the feature map size, and $k$ is the kernel size. Adding Laplacian noise is $O(n^2)$ which is not dominant.

    \item \textbf{Analyzer:} It contains A forward pass through the analyzer layer (linear/convolutional layer) with complexity $O(d)$, where $d$ is the feature map size with a backward pass for gradient updates which has compelxity $O(d)$.
    Thus, the total complexity is $ O(d)$ per epoch.

    \item \textbf{Adversary:} For $m_a$ adversary training steps per epoch, each step involves a forward pass through the adversary (linear/convolutional layer) with complexity $O(d)$, where $d$ is the feature map size and a backward pass for gradient updates that has complexity $O(d)$. Thus, the total complexity is $m_a \cdot O(d)$ per epoch.

    \item \textbf{Overall Complexity for Edge Pre-Training:}
    With $n_p$ epochs and an edge device with a computational speed of $s_e$, the total complexity is:
    \begin{equation}
    O(\frac{n_p}{s_e} \cdot (m \cdot n^2 \cdot k^2 + (m_a + 1) \cdot d)).
    \end{equation}
\end{itemize}

\subsection*{2. Edge-Cloud Training (Section~\ref{subsec:edge-cloud-train})}
\begin{itemize}
    \item \textbf{Feature Extraction:} 
    Similar to the edge feature extraction, forward and backward passes scale as:
    \begin{equation}
    O(m \cdot n^2 \cdot k^2 + m' \cdot n'^2 \cdot k'^2),
    \end{equation}
    where $m', n', k'$ are cloud feature extractor parameters (which exist inside the large cloud analyzer shown in Figure \ref{fig:3-steps}).
    \item \textbf{Analyzer:} It involves a forward pass through the analyzer decision making layer (linear layer) that has complexity $O(d')$, where $d'$ is the feature map size and a backward pass for gradient updates with complexity $O(d')$. Thus, the total complexity is $ O(d')$ per epoch.
    \item \textbf{Adversary:} Similar to the edge pre-training, the complexity is $m_a \cdot O(d)$ per epoch.

    \item \textbf{Overall Complexity for Edge-Cloud Training:}
    Including $n_t$ epochs and computational speed of $s_e$ and $s_c$ for the edge and cloud devices, we have:
    \begin{equation}
    O(\frac{n_t}{s_e} \cdot (m \cdot n^2 \cdot k^2 + m_a \cdot d) + \frac{n_t}{s_c} \cdot (m' \cdot n'^2 \cdot k'^2 + d' )).
    \end{equation}
\end{itemize}

\ 

\ 

\ 

\subsection*{3. Total Complexity}
Combining both stages, the total complexity is:

\begin{equation}
O(\frac{n_p}{s_e} \cdot (m \cdot n^2 \cdot k^2 + (m_a + 1) \cdot d)) \ + \end{equation} 
\begin{equation} O(\frac{n_t}{s_e} \cdot (m \cdot n^2 \cdot k^2 + m_a \cdot d) + \frac{n_t}{s_c} \cdot (m' \cdot n'^2 \cdot k'^2 + d' )). \nonumber
\end{equation}

\subsection*{4. Special Cases}
\begin{itemize}
    \item \textbf{Case 1 (Slow Edge Device):} If $s_e$ is small or, $n_p$ is large (slow edge device or too many edge pre-training epochs), edge pre-training dominates the complexity.
    \item \textbf{Case 2 (Large Cloud Models):} If $m'$, $n'$, $k'$, or $n_t$ grow (large cloud models or too many edge-cloud training epochs), edge-cloud training dominates the complexity.
\end{itemize}

\subsection*{5. Summary}
The PriPHiT method is designed to be computationally efficient, particularly for resource-constrained edge devices, by minimizing parameter counts on the edge. The overall complexity is influenced by key factors, including the computational speed of the edge device ($s_e$), model dimensions (parameters $m, n, k, m', n', k'$), and the number of adversarial or training steps. Achieving runtime efficiency requires a careful balance of the pre-training epochs ($n_p$), training epochs ($n_t$), and adversarial steps ($m_a$). In practical applications, as demonstrated in Section \ref{Sec:experiments}, the number of edge pre-training steps ($n_p$) is typically kept small to mitigate computational demands on resource-constrained edge devices.

\section{Figures of the Three Architectures Used in the Experiments \label{app:architecture}}

\begin{figure}[h!]
\centering
\includegraphics[scale=0.45]{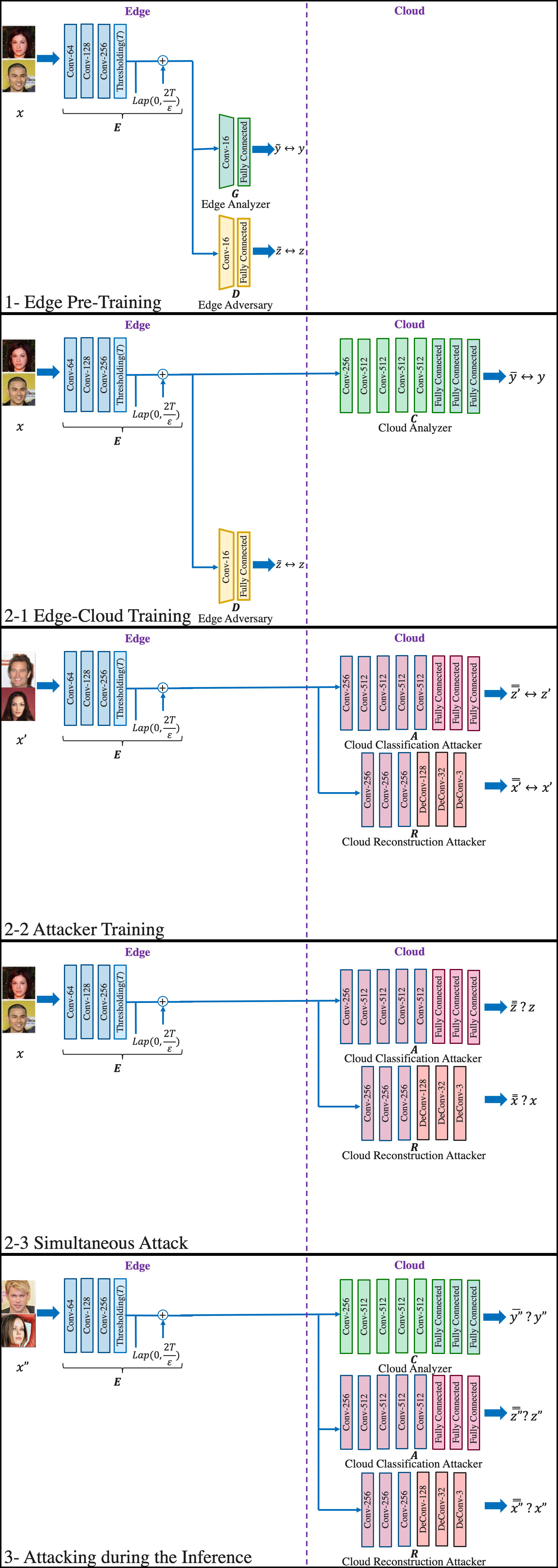}
\caption{The proposed method of privacy-preserving hierarchical training and its steps of edge-cloud execution when the VGG-11 architecture is used in the experiments. Architecture details are shown.}
\label{fig:vgg-11}
\end{figure}

\begin{figure}[h!]
\centering
\includegraphics[scale=0.45]{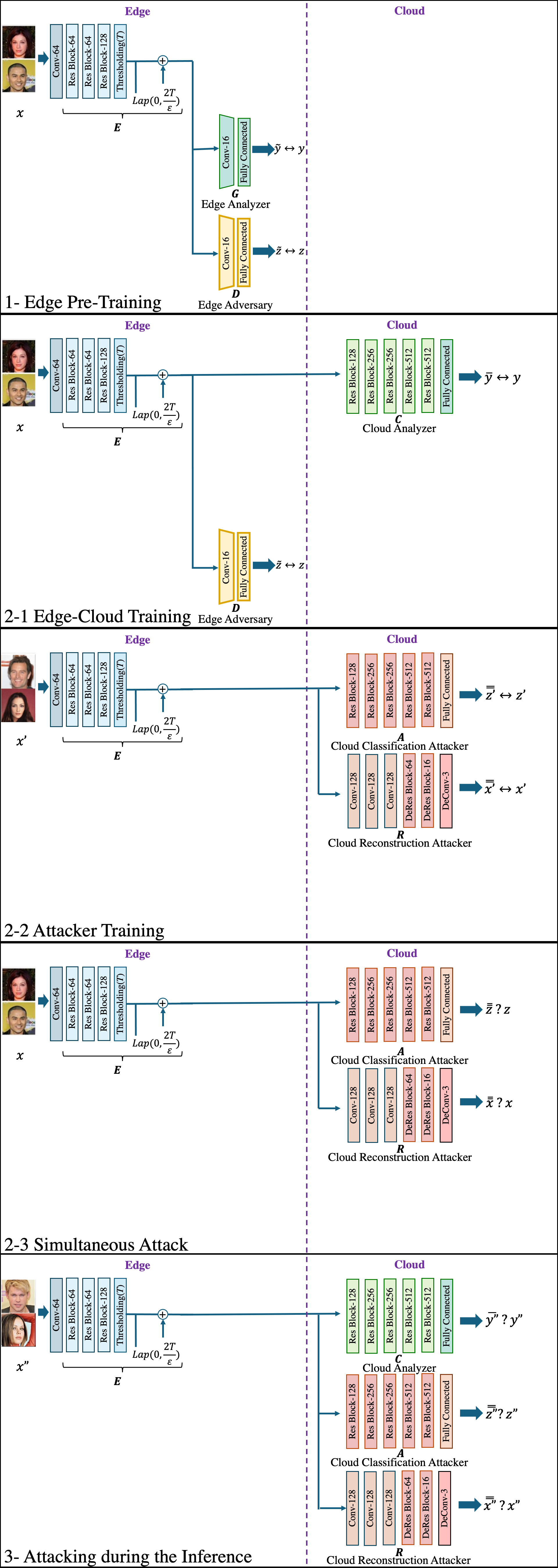}
\caption{The proposed method of privacy-preserving hierarchical training and its steps of edge-cloud execution when the ResNet-18 architecture is used in the experiments. Architecture details are shown.}
\label{fig:resnet-18}
\end{figure}

\begin{figure}[h!]
\centering
\includegraphics[scale=0.45]{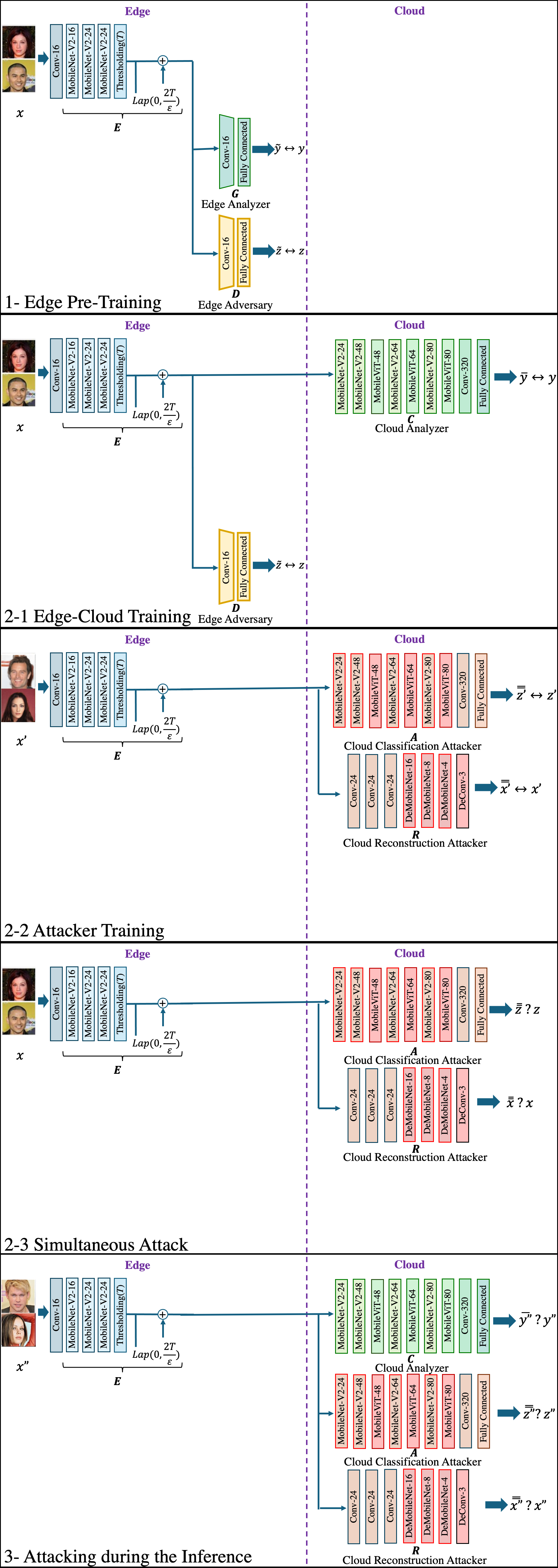}
\caption{The proposed method of privacy-preserving hierarchical training and its steps of edge-cloud execution when the MobileViT-xxs architecture is used in the experiments. Architecture details are shown.}
\label{fig:mobilevit-xxs}
\end{figure}

Figs.~\ref{fig:vgg-11}, \ref{fig:resnet-18}, and \ref{fig:mobilevit-xxs} show the details of the VGG-11, ResNet-18, and MobileViT-xxs architectures that have been used in our experiments in the different steps of PriPHiT execution.


\section{Results of the Simultaneous Attacks in Mouth Open versus Wearing Makeup Experiment \label{app:experiment}}

\begin{figure*}[h!]
\centering
\includegraphics[scale=0.5]{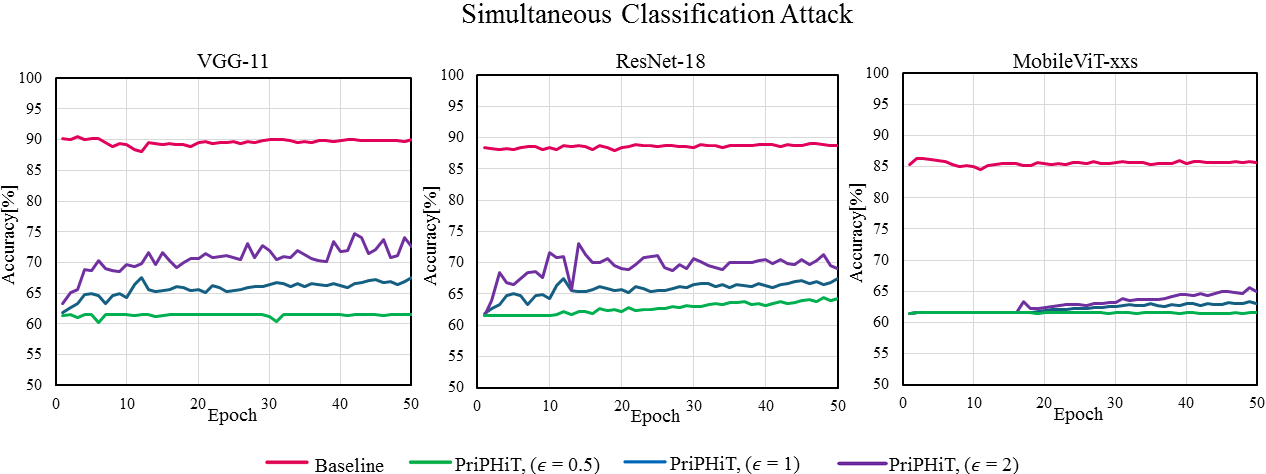}
\caption{The simultaneous classification attack accuracy during the training phase on the feature maps coming from the user's private data at each epoch for mouth open versus wearing makeup experiment on the CelebA dataset.}
\label{fig:simultaneous_attack_classmouth}
\end{figure*}

\begin{figure*}[h!]
\centering
\includegraphics[scale=0.5]{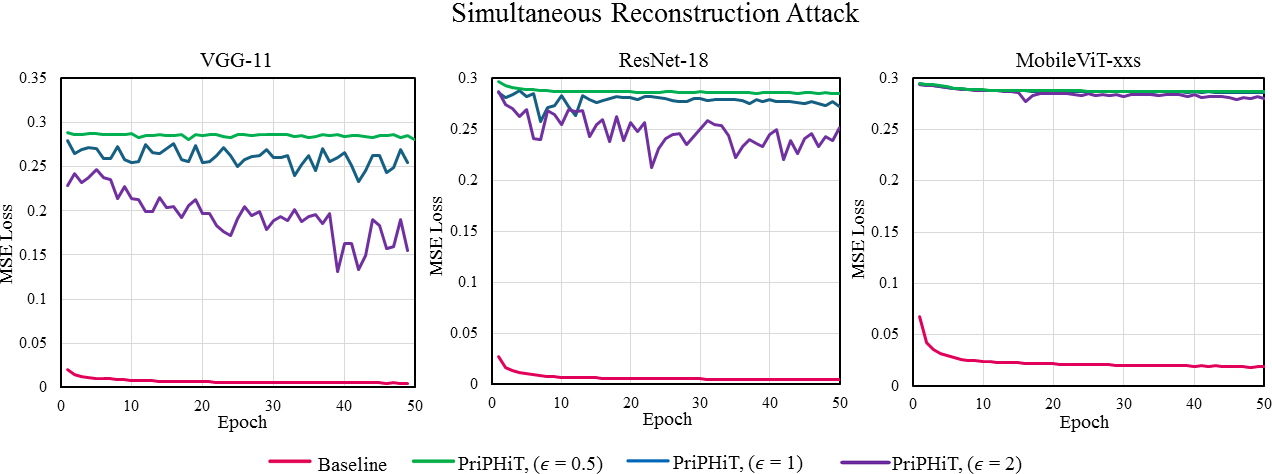}
\caption{MSE Loss of the simultaneous reconstruction attack on the feature maps coming from the user's private data compared with the user's original inputs at each epoch for mouth open versus wearing makeup experiment on the CelebA dataset.
}
\label{fig:simultaneous_attack_recmouth}
\end{figure*}

Figs.~\ref{fig:simultaneous_attack_classmouth} and ~\ref{fig:simultaneous_attack_recmouth} show the results of the simultaneous attacks during the training phase in the experiments when having an open mouth is selected as the desired content and wearing makeup is the sensitive content. Again, there is a substantial privacy leakage in the baselines as the simultaneous classification attack can provide a high accuracy and the deep reconstruction attack can provide a low MSE loss for the reconstructions in comparison to the original inputs of the user. However, in the PriPHiT method, the classification attack accuracies stay low and the MSE losses of the reconstruction attack stay high, indicating unsuccessful attacks. Additionally, the plots reach a plateau and cannot improve with more epochs, since the adversarial privacy-preserving training using the edge adversary is continuously done in this stage. A lower privacy budget ($\epsilon=0.5$) again provides a stronger preservation of privacy in comparison to a higher budget ($\epsilon=2$).

\section{White-Box Reconstruction Attack\label{app:white-box}}

In this section, we describe the details of the white-box reconstruction attacker that is used in the inference phase. We use a similar method as the one used in the recent work~\cite{guo2023mistnet}. In this attack, an image $u$ with the same size as the unseen input image $x$ is made step by step. It uses gradient decent on the sum of MSE loss between the extracted feature map of this image $E(u)$ and the available feature map of the input $x_{E}$, and the total variation loss $\text{TV}(u)$,
\begin{equation}
    u^* = \argminA_u  \left (E(u)-x_{E}\right)^2 + \alpha \text{ TV}(u)
\end{equation}
where $\alpha$ is a hyperparameter. In our experiments, this white-box reconstruction attacker performs $2000$ gradient decent steps on every feature map of the test set that is transmitted to the cloud and tries to reconstruct the inputs.

\section{Privacy Budget Analysis\label{app:budget}}
\begin{figure*}[h!]
\centering
\includegraphics[scale=0.58]{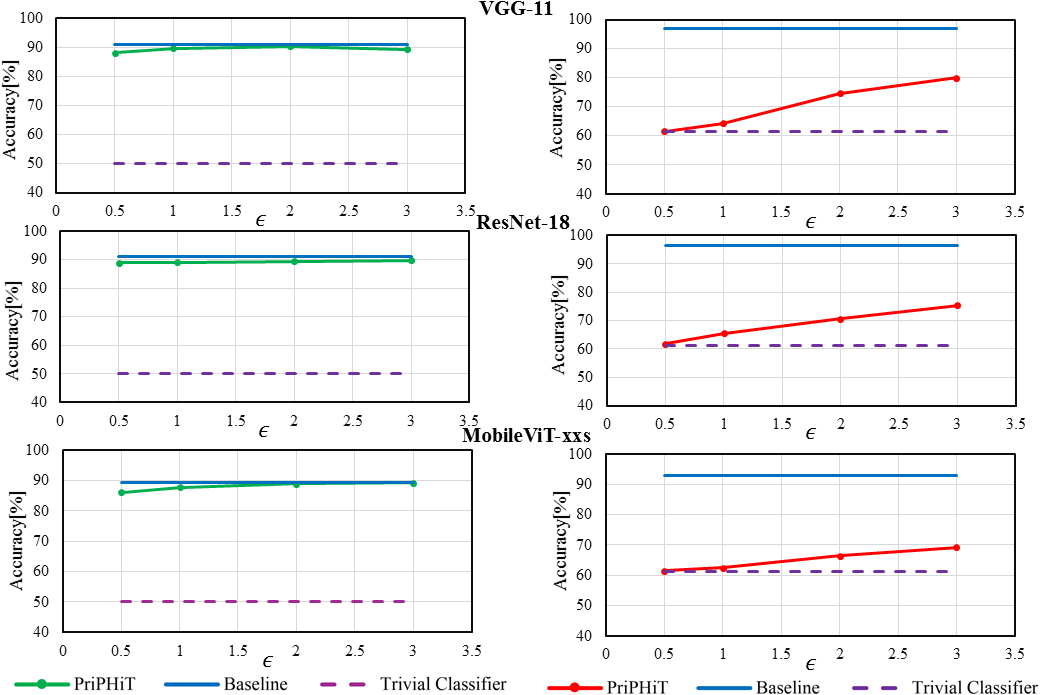}
\caption{The effect of privacy budget ($\epsilon$) on the accuracy of the analyzer versus the accuracy of the classification attacker in the inference phase when the desired attribute is smiling and the sensitive attribute is gender using different architectures on the CelebA dataset.}
\label{fig:epsilon_smiling}
\end{figure*}

\begin{figure*}[h!]
\centering
\includegraphics[scale=0.58]{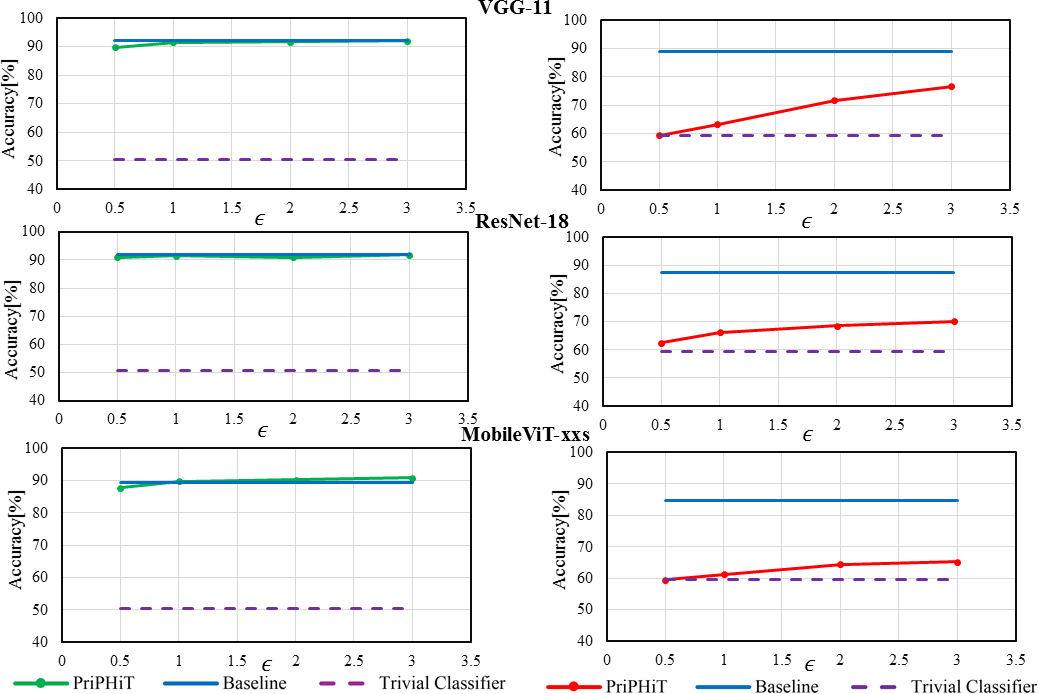}
\caption{The effect of privacy budget ($\epsilon$) on the accuracy of the analyzer versus the accuracy of the classification attacker in the inference phase when the desired attribute is having a slightly open mouth and the sensitive attribute is wearing makeup using different architectures on the CelebA dataset.}
\label{fig:epsilon_openmouth}
\end{figure*}

Figs.~\ref{fig:epsilon_smiling} and~\ref{fig:epsilon_openmouth} show the effect of the privacy budget $\epsilon$ on the accuracy of the cloud analyzer in comparison to the accuracy of the cloud classification attacker for the smiling versus gender and having an open mouth versus wearing makeup experiments on the CelebA dataset, respectively. In the PriPHiT method, reducing the privacy budget decreases the classification attacker's accuracy on the sensitive content significantly while causing a minimal cost in the analyzer accuracy on the desired content.

\

\section{Qualitative Results of the Reconstruction Attack in Mouth Open vs. Wearing Makeup Experiment \label{app:qualitative}}

\begin{figure*}[t!]
\centering
\includegraphics[scale=0.4]{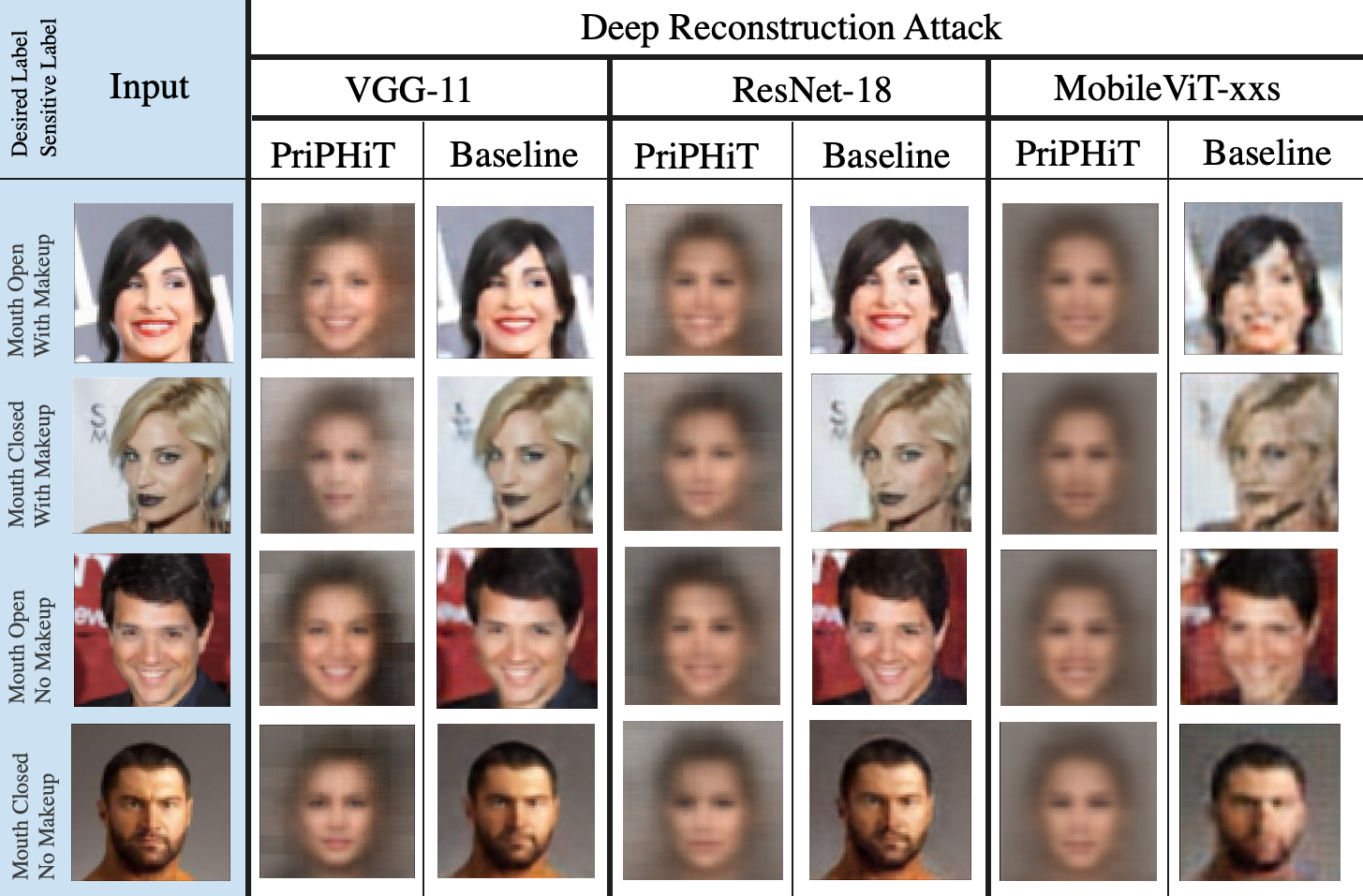}
\caption{Examples of the results of deep reconstruction attacks on PriPHiT ($\epsilon=1$) in comparison to the similar attacks on the baselines and the original unseen inputs from the testing subset of the CelebA dataset. Having an open mouth and wearing makeup are selected as the desired content and the sensitive content, respectively. Note the change of makeup indicators while keeping the desired content in the results of deep reconstruction attacks when the PriPHiT method is used.}
\label{fig:makeup_reconstruction_deep}
\end{figure*}

\begin{figure*}[t!]
\centering
\includegraphics[scale=0.4]{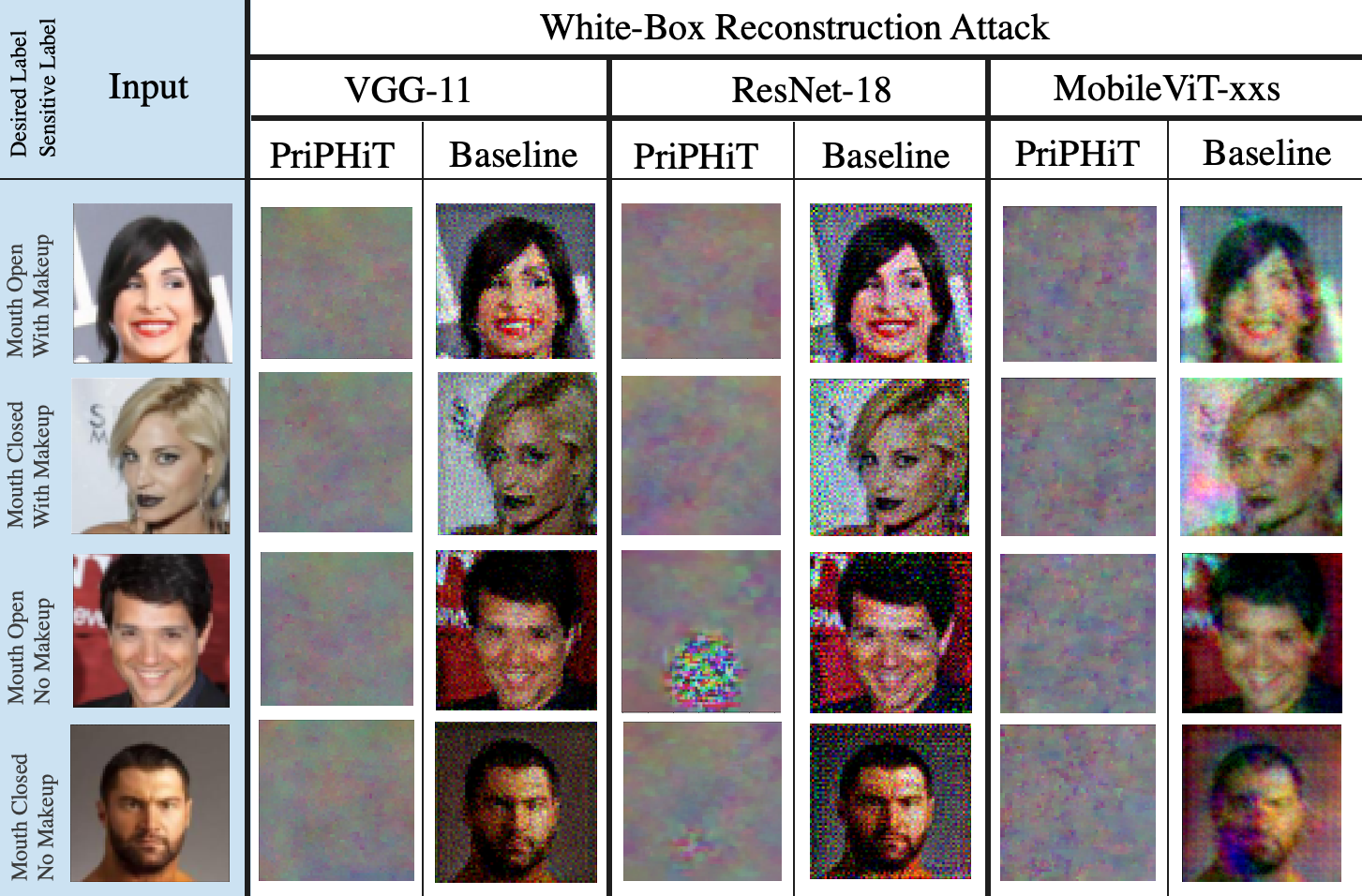}
\caption{Examples of the results of white-box reconstruction attacks on PriPHiT ($\epsilon=1$) in comparison to the similar attacks on the baselines and the original unseen inputs from the testing subset of the CelebA dataset. Having an open mouth and wearing makeup are selected as the desired content and the sensitive content, respectively. Note the failure of the attack when the PriPHiT method is used.}
\label{fig:makeup_reconstruction_white}
\end{figure*}

Figs.~\ref{fig:makeup_reconstruction_deep} and~\ref{fig:makeup_reconstruction_white} show examples of the results of the reconstruction attacks together with the unseen inputs from the testing subset for the mouth open versus wearing heavy makeup experiment on the CelebA dataset. We see that the output of both attacks can reconstruct the inputs with makeup indicators using the extracted feature maps from the baselines; however, they fail to reconstruct the inputs when the PriPHiT ($\epsilon=1$) method is used. It is interesting to see that in the outputs of the deep reconstruction attacks, the makeup indicators are changed, when PriPHiT is applied.


\end{document}